\documentclass[lettersize,journal]{IEEEtran}

\usepackage{algorithm}
\usepackage{algpseudocode}
\usepackage{float}

\usepackage{graphicx}
\usepackage{array}
\usepackage{caption}
\usepackage{etoolbox}
\usepackage[T1]{fontenc}
\usepackage{tabularx}
\usepackage{booktabs}
\usepackage{multirow}
\usepackage{amsmath}
\usepackage{siunitx}
\usepackage{enumitem}
\usepackage{threeparttable}
\usepackage{amsfonts}
\usepackage[caption=false,font=normalsize,labelfont=sf,textfont=sf]{subfig}
\usepackage{textcomp}
\usepackage{stfloats}
\usepackage{url}
\usepackage{verbatim}
\usepackage{cite}
\usepackage{xcolor}

\usepackage{booktabs}
\usepackage{multirow}
\usepackage{array}
\usepackage{tabularx}
\usepackage{makecell}

\newcommand{\imgwidth}{0.30\textwidth}

\newcommand{\includeimg}[2]{%
  \includegraphics[width=\linewidth]{./Image/#1/#2.png}%
}

\newcommand{\imgcell}[2]{%
  \begin{minipage}[c]{\imgwidth}
    \centering
    \includeimg{#1}{#2}\\[-2pt]
    {\scriptsize #2}
  \end{minipage}%
}

\begin{document}

\title{CityTrajBench: A Unified Benchmark for City-Scale Vehicle Trajectory Generation}

\author{Shibo~Zhu,
        Xiaodan~Shi{\IEEEauthorrefmark{1}},
        Dayin~Chen,
        Yuntian~Chen,~\IEEEmembership{Member,~IEEE,}
        Haoran~Zhang,~\IEEEmembership{Senior Member,~IEEE,}
        Tianhao~Wu{\IEEEauthorrefmark{1}},
        and Jinyue~Yan{\IEEEauthorrefmark{1}}

\thanks{\textcolor{black}{* Xiaodan Shi, Tianhao Wu and Jinyue Yan are the corresponding authors.}}
\thanks{Shibo Zhu and Dayin Chen are with Department of Building Environment and Energy Engineering, The Hong Kong Polytechnic University, Hong Kong SAR, China;
Eastern Institute for Advanced Study, Eastern Institute of Technology, Ningbo, China;
International Centre of Urban Energy Nexus, The Hong Kong Polytechnic University, Hong Kong SAR, China;
E-mail: shibo.zhu@connect.polyu.hk, 23038748r@connect.polyu.hk}
\thanks{\textcolor{black}{Xiaodan Shi is with Department of Computer and Systems Sciences, Stockholm University, Sweden. E-mail: xiaodan.shi@dsv.su.se}}
\thanks{\textcolor{black}{Yuntian Chen is with Zhejiang Key Laboratory of Industrial Intelligence and Digital Twin, Eastern Institute of Technology, Ningbo, China; Ningbo Institute of Digital Twin, Eastern Institute of Technology, Ningbo, China. E-mail: ychen@eitech.edu.cn}}
\thanks{\textcolor{black}{Tianhao Wu is with Eastern Institute for Advanced Study, Eastern Institute of Technology, Ningbo, Zhejiang, China; Zhejiang Key Laboratory of Industrial Intelligence and Digital Twin, Eastern Institute of Technology, Ningbo, China; Ningbo Key Laboratory of Advanced Manufacturing Simulation, Eastern Institute of Technology, Ningbo, China. E-mail: twu@eitech.edu.cn}}
\thanks{\textcolor{black}{Haoran Zhang is with LocationMind Inc., Tokyo 101-0042, Japan. E-mail: zhang\_ronan@locationmind.com}}
\thanks{Jinyue Yan is with Department of Building Environment and Energy Engineering, The Hong Kong Polytechnic University, Hong Kong SAR, China; International Centre of Urban Energy Nexus, The Hong Kong Polytechnic University, Hong Kong SAR, China. E-mail: j-jerry.yan@polyu.edu.hk}}

% {Zhu \MakeLowercase{\textit{et al.}}: CityTrajBench: A Unified Benchmark for City-Scale Vehicle Trajectory Generation}

\maketitle

\begin{abstract}
Urban trajectory generation is a fundamental task for transportation simulation, urban planning, and mobility analytics. However, systematic comparison across trajectory generation methods remains difficult because existing studies often rely on different datasets, preprocessing pipelines, trajectory representations, and evaluation metrics. This fragmentation makes it unclear whether reported performance differences arise from the generation mechanism itself or from inconsistent experimental protocols. To address this issue, we present \emph{CityTrajBench}, a unified benchmark framework and protocol for city-scale vehicle trajectory generation. CityTrajBench standardizes data ingestion, trajectory normalization, feature construction, model adaptation, map-aware post-processing, model selection, and multi-level evaluation under a common setting. It supports heterogeneous generators, including statistical baselines, VAE-based, GAN-based, diffusion-based, and flow-matching-based models, and evaluates them on three real-world urban trajectory datasets. The benchmark measures global spatial realism, trip-level distribution fidelity, trajectory-level geometric similarity, conditional mobility consistency, and efficiency. Experiments reveal clear trade-offs across model families: \emph{DiffTraj} is strongest on trajectory-level geometric fidelity, \emph{DiffRNTraj} is competitive on structure-sensitive global realism, and \emph{TrajFlow} provides a strong balance across realism, quality, conditional consistency, and efficiency. Meanwhile, a simple \emph{Markov} baseline remains competitive on coarse-grained trip and local-movement statistics. These findings show that urban trajectory generation quality is inherently multi-objective, that no single model dominates all criteria equally, and that CityTrajBench provides a reproducible benchmark protocol and testbed for future research on urban mobility generation.
\end{abstract}


\section{Introduction}

Urban trajectory generation aims to synthesize realistic movement sequences that reflect how vehicles or individuals travel in a city. It is an important capability for a wide range of applications, including transportation simulation, urban planning, mobility-demand analysis, digital twins, and privacy-preserving data sharing \cite{kong2023mobility, xiong2023trajsgan, li2024urban}. Compared with conventional prediction tasks, trajectory generation is more challenging because a useful generator must capture not only local movement dynamics, but also city-scale spatial coverage, trip-level statistics, and plausible route geometry. In practice, high-quality generation requires simultaneously preserving realism, diversity, contextual consistency, and structural validity \cite{zhu2024controltraj, zhu2023difftraj, wei2024diff}.

Driven by these demands, trajectory generation methods have evolved rapidly from classical probabilistic approaches to deep generative models. Early studies mainly relied on transition-based or statistical formulations, which are computationally lightweight and interpretable but limited in modeling complex long-range dependencies. More recent work has introduced variational autoencoders (VAEs), generative adversarial networks (GANs), diffusion models, and more recently flow-matching-style models to learn richer trajectory distributions directly from data \cite{chen2021trajvae, rao2020lstm, zhu2023difftraj, li2026trajflow}. In particular, diffusion-based methods have shown strong promise in modeling high-dimensional mobility patterns and incorporating controllability or road-network structure \cite{zhu2024controltraj, wei2024diff}. Despite this progress, a basic question remains insufficiently answered: \emph{which trajectory generation methods perform best, under what data conditions, and according to which notion of quality?}

\begin{figure*}[h]
\centering
\includegraphics[width=0.94\linewidth]{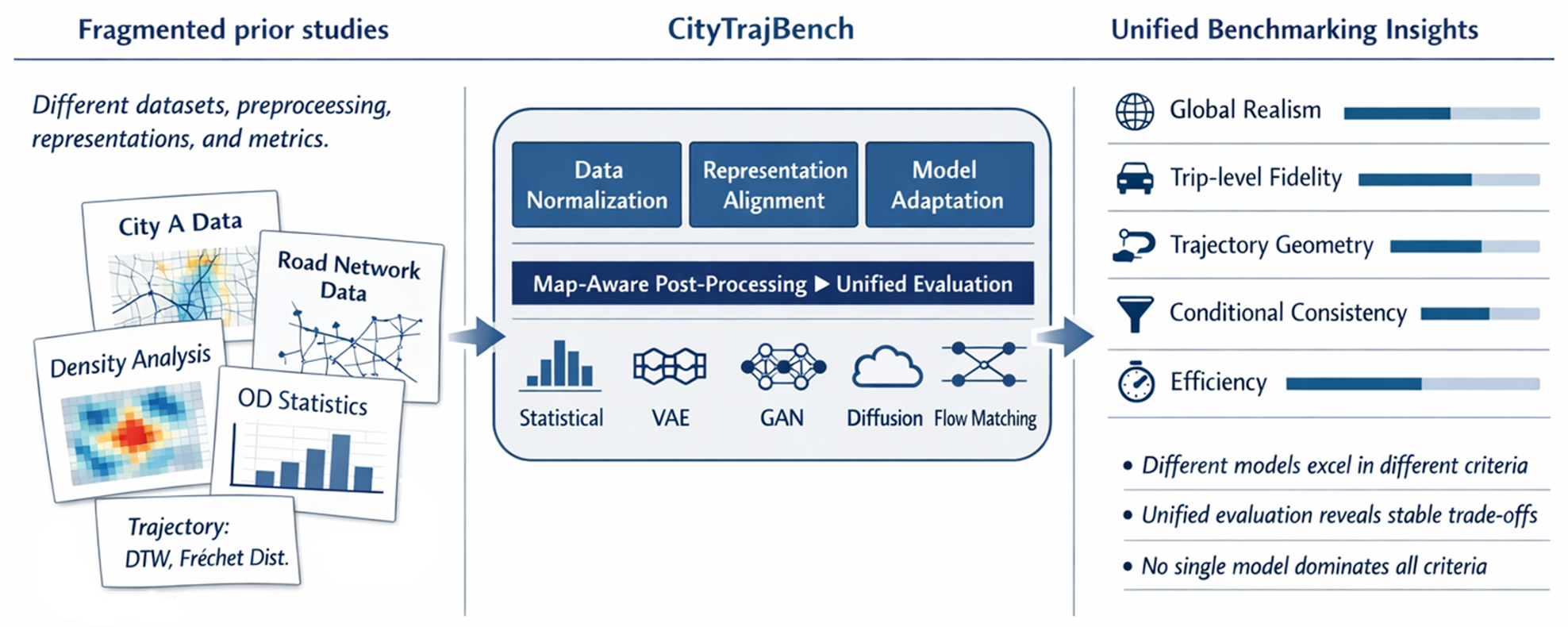}
% \caption{Introduction}
\caption{Motivation of CityTrajBench. Existing trajectory-generation studies often differ in datasets, preprocessing pipelines, trajectory representations, and evaluation metrics, making direct comparison difficult. CityTrajBench unifies benchmark-side normalization, representation alignment, model adaptation, map-aware post-processing, and multi-level evaluation to enable reproducible cross-model comparison.}
\label{fig:Introduction}
\end{figure*}

A major obstacle is the lack of a unified benchmark for city-scale vehicle trajectory generation. In contrast to trajectory prediction, where benchmark datasets and evaluation practices are relatively mature \cite{kothari2021human, feng2024unitraj}, generation studies are still commonly evaluated under incompatible settings. Existing work often differs in datasets, cleaning rules, temporal sampling, train--test splits, feature availability, trajectory representations, and metric choices. Some studies focus mainly on distributional similarity or visual inspection, whereas others emphasize trajectory-level distances or road-network validity \cite{chen2021trajvae, zhu2023difftraj, wei2024diff}. As a result, published results are difficult to compare directly, and conclusions about which model is ``better'' may depend as much on the evaluation protocol as on the generation method itself. Fig.~\ref{fig:Introduction} summarizes the central motivation of this work: prior trajectory-generation studies are often conducted under incompatible experimental settings, whereas CityTrajBench unifies benchmark-side preprocessing, representation alignment, model adaptation, and evaluation to enable fairer and more interpretable comparison.

This benchmarking problem is especially challenging in urban settings for three reasons. First, \emph{dataset heterogeneity} is substantial: different cities exhibit different road topologies, sampling frequencies, trip lengths, and mobility regimes, all of which can strongly affect model behavior. Second, \emph{representation heterogeneity} is unavoidable: some models generate directly in GPS space, while others operate in road-network, grid, or other constrained representations \cite{zhu2023difftraj, wei2024diff}. Without a unified preprocessing and post-processing protocol, these heterogeneous models cannot be compared fairly. Third, \emph{evaluation heterogeneity} remains unresolved: a model that matches city-wide density well may still generate unrealistic individual trajectories, while a model with low trajectory-level distance may fail to recover realistic origin--destination structure or hotspot coverage. A meaningful benchmark must therefore assess multiple aspects of generation quality simultaneously rather than relying on a single metric.

To address these challenges, we propose \emph{CityTrajBench}, a unified benchmark protocol for city-scale vehicle trajectory generation, implemented as a modular benchmark framework. CityTrajBench standardizes data ingestion, normalization, feature construction, model adaptation, map-aware post-processing, checkpoint selection, and evaluation under a common protocol. The framework is designed to support heterogeneous model families, including statistical, VAE-based, GAN-based, diffusion-based, and flow-matching-based generators, while reducing implementation-induced variability in cross-model comparison. Beyond protocol unification, CityTrajBench also provides a multi-perspective evaluation suite that jointly measures global spatial realism, trip-level distribution fidelity, trajectory-level geometric similarity, conditional mobility consistency, and computational efficiency.

Using CityTrajBench, we conduct a systematic empirical study on seven representative generators across three real-world urban trajectory datasets. The benchmark reveals several stable findings. \emph{DiffTraj} consistently achieves the best trajectory-level geometric fidelity, \emph{DiffRNTraj} is particularly strong on structure-sensitive global metrics such as density realism and hotspot recovery, and \emph{TrajFlow} offers a strong balance between global realism, trajectory-level quality, and computational efficiency. At the same time, even a simple \emph{Markov} baseline remains competitive on coarse-grained endpoint-related statistics, indicating that low-order transition structure still explains part of urban mobility regularity. These results show that trajectory generation quality is inherently multi-dimensional, and that no single model dominates all evaluation criteria equally.

The main contributions of this paper are summarized as follows:
\begin{itemize}
    \item We propose \emph{CityTrajBench}, a unified benchmark protocol and framework for city-scale vehicle trajectory generation, with standardized data preprocessing, model interfacing, map-aware post-processing, model selection, and reproducible evaluation.
    \item We establish a multi-level evaluation protocol that jointly assesses global spatial realism, trip-level statistical fidelity, trajectory-level geometric similarity, conditional origin--destination consistency, and computational efficiency.
    \item We provide a systematic benchmark study across multiple real-world datasets and heterogeneous model families, revealing stable trade-offs between macro-level realism, micro-level fidelity, structure-aware generation quality, and deployment cost.
\end{itemize}

The remainder of this paper is organized as follows. Section~II reviews related work. Section~III formulates the problem. Section~IV presents the CityTrajBench framework. Section~V describes the datasets, baselines, and experimental protocol. Section~VI reports the benchmark results and analyses. Section~VII concludes the paper.

\section{Related Work}

Research on trajectory generation has developed rapidly in recent years, spanning classical stochastic models, deep generative architectures, and road-network-aware generation methods. In parallel, the increasing complexity of trajectory data has also highlighted the need for principled evaluation and reproducible benchmarking. In this section, we review the literature from three perspectives: trajectory generation paradigms, road-network- and context-aware generation, and benchmarking and evaluation.

\subsection{Trajectory Generation Paradigms}

Early studies on mobility synthesis mainly relied on probabilistic and statistical formulations, such as Markov chains and related transition-based approaches \cite{gonzalez2008understanding, gambs2012next}. These methods are computationally lightweight and interpretable, and they can reproduce coarse-grained mobility regularities such as frequently visited regions or common origin--destination transitions. However, their expressive power is limited when modeling long-range spatial dependencies, complex route structures, and high-dimensional spatiotemporal variability in urban-scale trajectory data \cite{choi2021trajgail, zhu2023difftraj}.

With the development of deep generative modeling, learning-based approaches have become increasingly prominent. Variational autoencoder (VAE) frameworks provide a probabilistic latent-variable formulation for modeling trajectory distributions, enabling efficient sampling and smooth latent interpolation. They often produce stable training behavior and can preserve trajectory-level geometric similarity reasonably well \cite{chen2021trajvae}. However, they may over-smooth trajectories or under-represent rare mobility modes. Generative adversarial networks (GANs) have also been adopted for trajectory synthesis in both urban mobility generation and privacy-preserving synthetic trajectory publishing \cite{choi2021trajgail, rao2020lstm, liu2018trajgans}. Nevertheless, GAN-based models are often sensitive to training instability and mode collapse, which can affect both diversity and reproducibility.

More recently, diffusion models have emerged as a powerful alternative for trajectory generation. DiffTraj demonstrates that denoising diffusion probabilistic models can effectively synthesize realistic GPS trajectories \cite{zhu2023difftraj}, while subsequent studies such as ControlTraj \cite{zhu2024controltraj} and DiffRNTraj \cite{wei2024diff} incorporate controllability and road-network structure into the diffusion process. Flow-matching-based models have also recently shown promise in large-scale trajectory synthesis \cite{li2026trajflow}. Compared with adversarial methods, diffusion- and flow-based approaches typically offer stronger optimization stability and richer continuous trajectory modeling, but may still face challenges in efficiency, controllability, and protocol-sensitive evaluation.

\subsection{Road-Network- and Context-Aware Trajectory Generation}

A key challenge in urban trajectory generation is that mobility is strongly constrained by external factors, including road topology, departure time, trip distance, and destination semantics. To improve realism, a number of studies have incorporated contextual information or structural priors into generation models. Grid-based approaches discretize space into cells and model transitions or occupancy patterns over grids \cite{xu2021simulating, yuan2022activity}. These methods are convenient for large-scale learning, but they introduce a granularity--fidelity trade-off: coarse grids lose spatial precision, while fine grids increase sparsity and computational burden.

Other methods convert trajectories into image-like representations and then apply image-generation techniques \cite{ouyang2018non, cao2021generating}. Although such representations can leverage mature computer vision architectures, they may distort geometric continuity and complicate precise spatial evaluation. In contrast, road-network-aware models explicitly align generation with the graph structure of urban roads \cite{zhu2024controltraj}. DiffRNTraj \cite{wei2024diff}, for example, directly generates road-network-constrained trajectories through structure-aware diffusion, improving topological validity and corridor realism.

Despite these advances, existing models still differ substantially in their assumptions, input features, and output spaces. Some models generate in raw GPS space, whereas others operate on road edges or grid cells. Some support trip-level conditioning, while others are purely unconditional. This representation mismatch makes fair empirical comparison difficult and motivates a benchmark framework with unified preprocessing, post-processing, and evaluation interfaces \cite{feng2024unitraj, kothari2021human}.

\subsection{Benchmarking and Evaluation of Trajectory Models}

Although benchmarking has become standard in many machine learning domains, unified evaluation for trajectory generation remains limited. Existing benchmarks are more common in trajectory prediction than in trajectory generation \cite{feng2024unitraj, kothari2021human}. For example, prior benchmark efforts often focus on forecasting short-term future motion, pedestrian trajectory prediction, or downstream tasks such as traffic prediction and anomaly detection. These settings are informative, but they do not fully address the challenges of unconditional or conditional generation of complete urban trajectories.

Current evaluation practice in trajectory generation is highly fragmented. Different studies use different datasets, preprocessing rules, coordinate systems, temporal resolutions, and train/test splits. In addition, evaluation metrics are often inconsistent across papers \cite{feng2024unitraj, kothari2021human, chen2021trajvae}. Some focus on distributional similarity, such as density matching or origin--destination statistics \cite{zhu2023difftraj}; others report trajectory-level distances such as Dynamic Time Warping (DTW) or Fr\'echet distance \cite{li2026trajflow}; still others rely heavily on qualitative visualization. As a result, the literature lacks a common basis for answering whether one generation method is broadly better than another, or whether different methods are simply optimized for different notions of quality.

This issue is especially pronounced for city-scale vehicle trajectory generation. A model may achieve good trajectory-level similarity yet fail to reproduce global urban coverage. Conversely, a model may match city-wide density well but produce implausible individual paths. Moreover, GPS-based and road-network-based models are often evaluated under different assumptions, which further weakens comparability. A useful benchmark should therefore satisfy at least four requirements: (1) support multiple datasets with standardized preprocessing; (2) compare heterogeneous model families under a common protocol; (3) evaluate both macro-level and micro-level generation quality; and (4) provide reproducible implementations and evaluation scripts. CityTrajBench is designed to meet these requirements for city-scale vehicle trajectory generation.

\section{Problem Formulation}

In this section, we formalize the city-scale vehicle trajectory generation problem considered in this benchmark. Unlike trajectory prediction, which estimates future motion conditioned on observed historical trajectories \cite{kothari2021human, feng2024unitraj}, trajectory generation aims to synthesize complete trajectories that are statistically and geometrically consistent with real urban mobility data. In the context of CityTrajBench, the goal is not only to generate plausible individual trajectories, but also to evaluate whether different generative models can faithfully reproduce mobility patterns under a unified protocol across datasets, representations, and metrics.

\subsection{Trajectory Representation}

Let a trajectory be denoted by
\begin{equation}
T = \left\{ p_t \right\}_{t=1}^{n},
\qquad
% p_t = (\lambda_t, \phi_t),
p_t = (\phi_t,  \lambda_t),
\end{equation}
where \(p_t\) is the spatial position of a vehicle at step \(t\),  \(\phi_t\) denotes latitude, \(\lambda_t\) denotes longitude,  and \(n\) is the trajectory length (i.e., the number of sampled points before fixed-length normalization). Accordingly, a trajectory is an ordered polyline in geographic space that describes the movement of one trip from its origin to its destination.

Depending on the model family, a trajectory may also admit an alternative road-network-aligned representation. Let the road network be modeled as a graph
\begin{equation}
G = (\mathcal{V}, \mathcal{E}),
\end{equation}
where \(\mathcal{V}\) and \(\mathcal{E}\) denote the node and edge sets, respectively. After map matching, a trajectory can be represented as
\begin{equation}
T^{(\mathrm{rn})} = \left\{ (e_t, \rho_t) \right\}_{t=1}^{n},
\end{equation}
where \(e_t \in \mathcal{E}\) is the matched road segment and \(\rho_t \in [0,1]\) denotes the relative progress along that segment. This dual representation allows CityTrajBench to compare GPS-based and road-network-based generators within a common benchmarking framework.

\subsection{Contextual Features}

In addition to the trajectory itself, a generator may condition on auxiliary trip descriptors or structural information. For a trip \(T_i\), we denote its optional conditioning features by \(F_i\). In CityTrajBench, these features can include:
\begin{itemize}
    \item \textbf{Trip-level descriptors}, such as departure-time slot, travel distance, travel duration, trajectory length, average step distance, average speed, and discretized start/end region identifiers.
    \item \textbf{Road-network features}, such as road embeddings and connectivity constraints.
\end{itemize}

These contextual signals can be used to support either unconditional generation, where trajectories are sampled only from the learned mobility distribution, or conditional generation, where the generated trajectory is guided by trip-level or network-level descriptors.

\subsection{Generation Task Formulation}

Let
\begin{equation}
\mathcal{D} = \left\{ (T_i, F_i) \right\}_{i=1}^{N}
\end{equation}
denote a dataset of \(N\) real trajectories and their associated optional features. The objective of trajectory generation is to learn a generative model \(G_\theta\), parameterized by \(\theta\), that approximates the data distribution over trajectories.

For the unconditional setting, the model aims to learn
\begin{equation}
p_\theta(T) \approx p_{\mathrm{data}}(T),
\end{equation}
and generates a synthetic trajectory
\begin{equation}
\widehat{T} \sim p_\theta(T).
\end{equation}

For the conditional setting, the model aims to learn
\begin{equation}
p_\theta(T \mid F) \approx p_{\mathrm{data}}(T \mid F),
\end{equation}
so that, given a feature vector \(F\) and possibly a latent noise variable \(z\), it produces
\begin{equation}
\widehat{T} = G_\theta(z, F).
\end{equation}

The output \(\widehat{T}\) should preserve the essential characteristics of real urban mobility data, including spatial realism, trip-level statistical consistency, and plausible trajectory geometry. For road-network-aware methods, the generated trajectory should additionally satisfy topological consistency with the underlying road graph.

\subsection{Benchmark Objective}

From a benchmarking perspective, the problem is not limited to learning a single high-performing generator. Instead, the central objective is to evaluate heterogeneous trajectory generation methods under a common and reproducible protocol. Let
\begin{equation}
\mathcal{M} = \{M_1, M_2, \dots, M_K\}
\end{equation}
be a set of trajectory generators with different modeling assumptions and output spaces. For each model \(M_k\), CityTrajBench trains the model on the same training split, generates a set of synthetic trajectories \(\widehat{\mathcal{D}}_k\), and evaluates it against the real test distribution \(\mathcal{D}^{\mathrm{test}}\) using a unified metric suite:
\begin{equation}
\mathbf{m}_k
=
\mathrm{Eval}\!\left(\widehat{\mathcal{D}}_k, \mathcal{D}^{\mathrm{test}}\right)
\in \mathbb{R}^{d}.
\end{equation}

Here, \(\mathbf{m}_k\) is a \(d\)-dimensional metric vector that jointly characterizes generation quality from multiple perspectives, including global spatial fidelity, trip-level distribution matching, trajectory-level geometric similarity, conditional consistency, and efficiency. The benchmark is therefore designed to compare models fairly across heterogeneous representations while exposing trade-offs that would be obscured by single-metric evaluation.

\subsection{Evaluation Criteria and Challenges}

A practically useful urban trajectory generator should satisfy several requirements simultaneously:
\begin{itemize}
    \item \textbf{Global realism}: the generated trajectory set should reproduce city-scale spatial density and major mobility corridors.
    \item \textbf{Trip-level fidelity}: the generated trips should match the distributions of endpoints, travel lengths, and local movement increments.
    \item \textbf{Trajectory-level plausibility}: individual generated trajectories should exhibit realistic geometric shapes and route structures.
    \item \textbf{Conditional consistency}: when origin, destination, departure time, or other descriptors are provided, the generated trajectories should remain coherent with these conditions.
    % \item \textbf{Structural validity}: for road-network-constrained settings, generated trajectories should respect connectivity and topology constraints.
\end{itemize}

These requirements are often competing in practice. A model may match global density distributions well while producing unrealistic individual paths, or achieve low trajectory-level distance while failing to recover realistic city-wide coverage. This multi-objective nature of the problem motivates the design of CityTrajBench: a unified benchmark that evaluates trajectory generation comprehensively rather than through a single notion of quality.

\begin{figure*}[h]
\centering
\includegraphics[width=\linewidth]{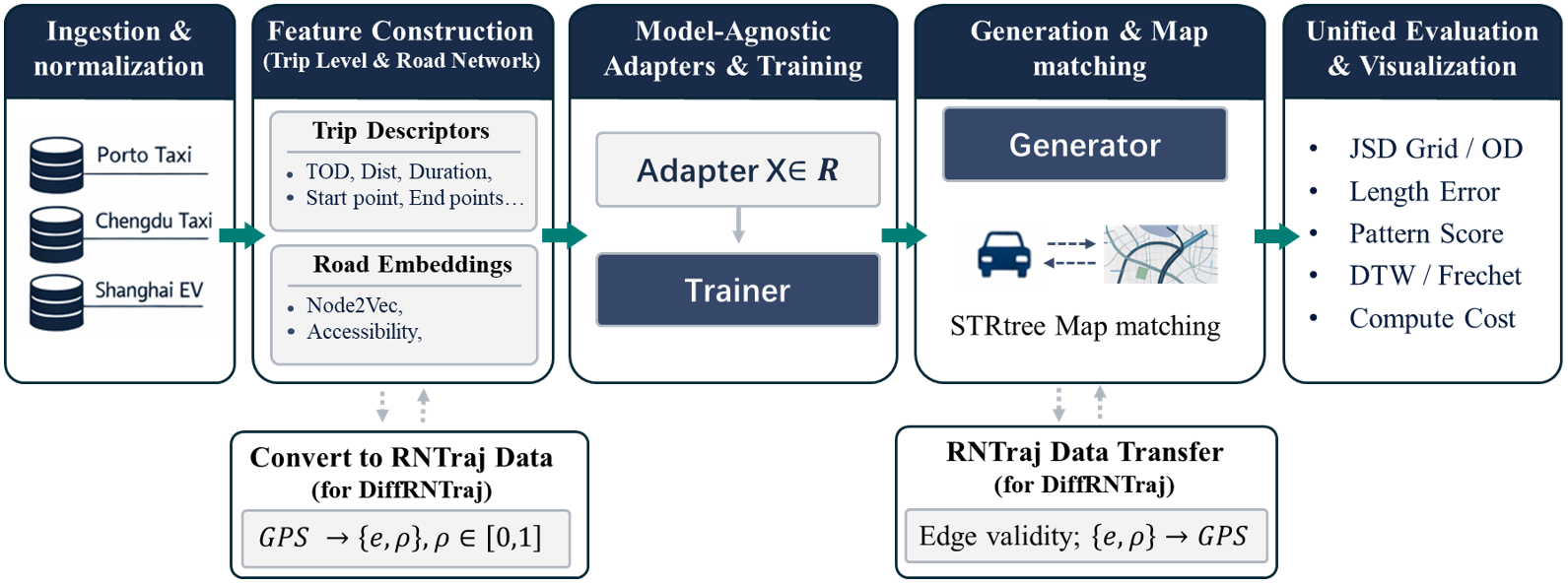}
\caption{Overview of the CityTrajBench framework. CityTrajBench standardizes benchmark-side data ingestion and normalization, representation alignment, feature construction, model adaptation and training, trajectory generation, map-aware post-processing, and unified evaluation and visualization across heterogeneous trajectory generators.}
\label{fig:Framework}
\end{figure*}

\section{CityTrajBench}
\label{sec:method}
The goal of this section is not to introduce a new trajectory generator, but to define the benchmark-side protocol that enables fair comparison across heterogeneous generators.
We present \emph{CityTrajBench}, a unified benchmark framework and protocol for city-scale vehicle trajectory generation. CityTrajBench organizes benchmarking into six benchmark-side stages: (i) data ingestion and normalization, (ii) representation alignment and feature construction, (iii) model-specific adaptation and training, (iv) trajectory generation, (v) map-aware post-processing, and (vi) unified evaluation and visualization. Fig.~\ref{fig:Framework} presents an overview of these modules and their interactions.

CityTrajBench is designed around three benchmark principles. \emph{Fidelity} requires that preprocessing and evaluation preserve the geographic and structural characteristics of real urban mobility data. \emph{Comparability} requires that dataset splits, representation conversion, model selection rules, and metric implementations be standardized at the benchmark level. \emph{Extensibility} requires that the framework support heterogeneous model families, including statistical, VAE-based, GAN-based, diffusion-based, and flow-matching-based generators, without forcing them into an artificial one-size-fits-all interface. Together, these principles define the methodological basis of CityTrajBench.

\subsection{System Overview}
\label{subsec:overview}

Let $\mathcal{D}=\{(T_i,F_i)\}_{i=1}^{N}$ denote a dataset of trips, where $T_i=\{(\phi_t,\lambda_t)\}_{t=1}^{n_i}$ is a GPS trajectory and $F_i$ is an optional descriptor vector. CityTrajBench exposes the following benchmark-level interfaces:
\begin{align}
\texttt{Reader:}&\quad \mathcal{S}\rightarrow \mathcal{D}, \\
\texttt{Preprocess:}&\quad \mathcal{D}\rightarrow \widetilde{\mathcal{D}}, \\
\texttt{Adapter:}&\quad (\widetilde{\mathcal{D}},\Pi)\rightarrow \mathcal{X}, \\
\texttt{Trainer:}&\quad \mathcal{X}\rightarrow \Theta, \\
\texttt{Generator:}&\quad (\Theta,F)\mapsto \widehat{\mathcal{D}}, \\
\texttt{Evaluator:}&\quad (\widehat{\mathcal{D}},\mathcal{D}^{\mathrm{test}})\mapsto \mathbb{R}^{K},
\end{align}
where $\mathcal{S}$ is a storage descriptor, $\widetilde{\mathcal{D}}$ is the normalized dataset, $\Pi$ denotes model-specific input contracts, $\mathcal{X}$ denotes model-ready tensors, $\Theta$ learned parameters, and $\mathbb{R}^{K}$ the benchmark metric vector.

These abstractions separate benchmark-side protocol design from model-side implementation. In particular, CityTrajBench standardizes preprocessing, representation conversion, evaluation, and reporting, while allowing different generators to retain their original modeling assumptions and input/output spaces as much as possible.

\subsection{Data Ingestion and Normalization}
\label{subsec:ingest}

To support fair comparison across heterogeneous datasets and model families, all data are processed under the same CityTrajBench pipeline. Trajectories are split at the trajectory level by \texttt{moveid}, so that all points from the same trip remain in the same subset. Unique \texttt{moveid}s are randomly shuffled and partitioned into training, validation, and test subsets with a ratio of 70\%/15\%/15\%.

After cleaning and dataset-specific normalization, each trajectory is further converted to a benchmark-standard sequence representation with a fixed length of \(L=200\). Trajectories shorter than \(L\) points are linearly interpolated to length \(L\), while longer trajectories are truncated to their first \(L\) points. The temporal order of points within each trajectory is preserved throughout preprocessing. This fixed-length normalization reduces variability caused by heterogeneous sampling density and sequence duration, and helps ensure comparable model inputs across datasets and baseline families. At the same time, it should be noted that this benchmark-side standardization also changes the raw sequence representation: interpolation may smooth short trajectories, while prefix truncation may discard part of long-trip geometry. Accordingly, the reported benchmark results should be interpreted as generation quality under a controlled fixed-horizon representation rather than as a direct evaluation of the original raw variable-length trajectories. Latitude and longitude are then linearly normalized to \([-1,1]\) using the dataset-specific geographic bounding box.

Trip-level descriptors are constructed from the normalized preprocessing pipeline, including departure-time slot, trip distance, trip duration, trajectory length, average step distance, average speed, and discretized start/end region identifiers. In CityTrajBench, trajectories are represented under a unified 30-second sampling setting. Accordingly, duration- and speed-related descriptors are computed consistently from the benchmark-standardized trajectory representation. Absolute departure-time annotation, however, is not uniformly available across datasets: for datasets with valid timestamp information such as Shanghai, departure-time is encoded as one of 288 five-minute slots within a day, whereas for datasets such as Porto and Chengdu, where the processed data do not retain real departure timestamps, the departure-time feature is set to zero as a placeholder. For road-network-aware methods, trajectories are further aligned with the road graph and converted into network-constrained representations. Unless otherwise stated, dataset-level spatial partitions, histogram-construction rules, and trajectory-matching settings are fixed per dataset and shared across all compared methods.

The exact benchmark configuration used in this study, including split rules, spatial bounds, sequence horizon, and evaluation grids, is summarized in Appendix~\ref{app:Configuration_summary}.

\subsection{Road-Network Representation Alignment and Map Matching}
\label{subsec:mapmatch}

A central difficulty in benchmarking urban trajectory generators is representation heterogeneity: some models operate directly in GPS space, whereas others require road-network-aligned inputs or produce road-constrained outputs. CityTrajBench addresses this issue through a reproducible representation-alignment module based on deterministic geometric projection. The purpose of this module is benchmark-side conversion and evaluation consistency rather than state-of-the-art map matching.

Let the road network be represented by $G=(\mathcal{V},\mathcal{E})$. For a candidate road polyline segment $\mathsf{seg}=\{(\phi_k,\lambda_k)\}_{k=1}^{m}$, we consider a curve parameterization $\gamma:[0,1]\rightarrow\mathbb{R}^{2}$ and project a GPS point $p=(\phi,\lambda)$ onto its closest point on the segment:
\begin{equation}
s^\star \in \arg\min_{s\in[0,1]} d_{\mathrm{geo}}\bigl(p,\gamma(s)\bigr),
\end{equation}
where $d_{\mathrm{geo}}(\cdot,\cdot)$ denotes geodesic distance. We then compute the along-segment arclength up to the projected point,
\begin{equation}
\ell^\star
=
\int_{0}^{s^\star}\|\gamma'(u)\|\,du,
\end{equation}
and normalize it by the segment length $L(\mathsf{seg})$ to obtain a relative progress variable
\begin{equation}
\rho=\ell^\star/L(\mathsf{seg})\in[0,1].
\end{equation}
The matched segment identifier is denoted by $e^\star\in\mathcal{E}$.

In practice, CityTrajBench constructs an \texttt{STRtree} over road polylines for candidate retrieval, queries a small set of nearby candidate segments for each GPS point, and assigns the point to the nearest segment by geometric projection. 
Repeating this process for all points in a trajectory yields an edge-aligned sequence $\{(e_t^\star,\rho_t)\}_{t=1}^{n}$ that preserves the original point order and benchmark sampling alignment.

We emphasize that this module is used primarily as a deterministic representation-conversion layer. It provides a common edge-aligned surrogate representation for preprocessing and post-generation conversion, and is supplemented by lightweight validity checks described in Sec.~\ref{subsec:post}. It should therefore be interpreted as a benchmark-consistency mechanism rather than as a claim of optimal map-matching accuracy.

\textbf{Complexity}
Candidate retrieval through the spatial index is $O(\log |\mathcal{E}|)$ per point on average, while local interpolation is linear in the number of vertices of the retrieved segment. The overall complexity is therefore near-linear in the number of GPS samples.

\subsection{Feature Construction}
\label{subsec:features}

CityTrajBench supports multi-granular feature extraction at the benchmark level. These features are intended to standardize feature availability and documentation, not to force identical feature usage across all baselines.

The benchmark can expose two broad classes of descriptors:
\begin{itemize}
\item \textbf{Trip-level descriptors} $F^{(\mathrm{trip})}$, including departure-time slot, trip distance, trip duration, trajectory length, average step distance, average speed, and discretized start/end region identifiers.
\item \textbf{Road-network descriptors} $F^{(\mathrm{rn})}$, including road embeddings and connectivity constraints
\end{itemize}

Depending on the baseline architecture, these descriptors may be concatenated with trajectory inputs, routed through model-specific conditioning branches, or not used at all. To avoid structural unfairness, CityTrajBench explicitly distinguishes between \emph{feature availability} at the benchmark level and \emph{feature usage} at the model level. The latter is documented separately for each baseline in the experimental section. This design improves transparency in cross-model comparison while avoiding an artificial one-size-fits-all input interface across fundamentally different model families.

\subsection{Model-Agnostic Adapters and Training}
\label{subsec:adapters}

To support heterogeneous generators under a common protocol, CityTrajBench uses model-specific adapters to convert benchmark-standardized trajectories and descriptors into model-ready tensors:
\begin{equation}
\mathrm{Adapt}(\widetilde{\mathcal{D}},F)\rightarrow X\in\mathbb{R}^{N\times L\times D},
\end{equation}
where $N$ is the number of samples, $L$ is the sequence horizon used by the target model interface, and $D$ is the per-step feature dimension. Depending on the baseline, $X$ may contain GPS coordinates, road-network embeddings, temporal encodings, or other supported inputs.

This adapter layer is intentionally thin: it preserves the original input assumptions of each model as much as possible while ensuring benchmark-level consistency in data preparation and reporting. All models are trained on the same train/validation split of a given dataset. Where original implementations expose compatible training settings, they are retained; otherwise, benchmark-default settings are used to reduce undocumented reproduction variance.

For reproducibility, each model is evaluated under three random seeds (42, 52, and 62), and benchmark tables report mean$\pm$std over these repeated runs. This practice reduces protocol variance and makes observed differences more attributable to the generation mechanism rather than accidental training randomness.

\subsection{Map-aware Post-processing}
\label{subsec:post}
Because different generators produce outputs in different spaces, CityTrajBench converts all generated samples into a common evaluation representation before metric computation. For GPS-space generators, normalized outputs are first mapped back to dataset-specific geographic coordinates. For road-network-aware generators such as DiffRNTraj, outputs are treated as already represented in GPS/road-network space. Generated trajectories are then post-processed by point-wise road-network map matching: each generated point is projected onto its nearest road polyline using the same benchmark-side spatial index and geodesic projection rule as in Sec.~\ref{subsec:mapmatch}. This post-processing step is used to improve evaluation consistency across heterogeneous output spaces rather than to perform full route reconstruction or global path optimization. Consequently, the reported realism metrics should be interpreted as quality after benchmark-side representation unification and lightweight map-aware projection, not as a pure evaluation of raw decoder outputs before any benchmark-side conversion.

\subsection{Unified Evaluation Protocol}
\label{subsec:metrics}
To ensure fair and reproducible comparison, all metrics are computed under a unified benchmark protocol. Let the real trajectory set be $\mathcal{D}=\{T_i\}_{i=1}^{N_r}$ and the generated set be $\widehat{\mathcal{D}}=\{\widehat{T}_j\}_{j=1}^{N_g}$. Before evaluation, all trajectories are converted to a common geographic representation and mapped back to the original coordinate range of the corresponding dataset. Unless otherwise stated, city-level distributions are computed using the same geographic bounds and the same spatial partition for both real and generated data.

CityTrajBench evaluates five complementary dimensions: \emph{global spatial realism}, \emph{trip-level distribution fidelity}, \emph{trajectory-level geometric similarity}, \emph{conditional mobility consistency}, and \emph{computational efficiency}. The purpose of this metric suite is not to impose a single ranking criterion, but to expose trade-offs between different notions of generation quality.

\subsubsection{Global spatial realism.}
We first assess whether a generated set reproduces city-scale density structure and major mobility hotspots.

\textbf{Density Error.}
The city bounding box is partitioned into a fixed $G\times G$ grid, shared by the real and generated sets on the same dataset. Let $p^{\mathrm{dens}}$ and $\hat{p}^{\mathrm{dens}}$ denote the corresponding normalized cell-density distributions. We define
\begin{equation}
\mathrm{DensityError}
=
\mathrm{JSD}\!\left(p^{\mathrm{dens}},\hat{p}^{\mathrm{dens}}\right),
\end{equation}
where
\begin{equation}
\mathrm{JSD}(p,q)
=
\frac{1}{2}\mathrm{KL}(p\|m)+\frac{1}{2}\mathrm{KL}(q\|m),
~ m=\frac{1}{2}(p+q).
\end{equation}

\textbf{Hotspot Pattern Score.}
Let $\Omega_K(p^{\mathrm{dens}})$ and $\Omega_K(\hat{p}^{\mathrm{dens}})$ denote the top-$K$ densest cells in the real and generated distributions. We compute
\begin{equation}
\mathrm{Precision}_K
=
\frac{
\left|
\Omega_K(p^{\mathrm{dens}})
\cap
\Omega_K(\hat{p}^{\mathrm{dens}})
\right|
}{
\left|
\Omega_K(\hat{p}^{\mathrm{dens}})
\right|
},
\end{equation}
\begin{equation}
\mathrm{Recall}_K
=
\frac{
\left|
\Omega_K(p^{\mathrm{dens}})
\cap
\Omega_K(\hat{p}^{\mathrm{dens}})
\right|
}{
\left|
\Omega_K(p^{\mathrm{dens}})
\right|
},
\end{equation}
and define
\begin{equation}
\mathrm{PatternScore}
=
\frac{
2\cdot \mathrm{Precision}_K \cdot \mathrm{Recall}_K
}{
\mathrm{Precision}_K+\mathrm{Recall}_K
}.
\end{equation}
This metric emphasizes recovery of dominant spatial hotspots rather than full-density calibration alone.

\subsubsection{Trip-level distribution fidelity.}
We next evaluate whether generated trips preserve coarse-grained mobility statistics.

\textbf{Trip Error.}
Let $p^{\mathrm{start}},p^{\mathrm{end}}$ be the empirical start and end distributions of the real set, and let $\hat{p}^{\mathrm{start}},\hat{p}^{\mathrm{end}}$ be their generated counterparts. We define
\begin{equation}
\mathrm{TripError}
=
\frac{1}{2}
\left[
\mathrm{JSD}\!\left(p^{\mathrm{start}},\hat{p}^{\mathrm{start}}\right)
+
\mathrm{JSD}\!\left(p^{\mathrm{end}},\hat{p}^{\mathrm{end}}\right)
\right]
\end{equation}

\textbf{Length Error.}
For each trajectory $T$, we compute the total travel length
\begin{equation}
\ell(T)=\sum_{t=2}^{L} d_{\mathrm{geo}}\!\big((\phi_{t-1},\lambda_{t-1}),(\phi_t,\lambda_t)\big).
\end{equation}
Using shared histogram bins constructed from pooled real and generated samples on the same dataset, we compute
\begin{equation}
\mathrm{LengthError}
=
\mathrm{JSD}\!\left(p^{\ell},\hat{p}^{\ell}\right),
\end{equation}
where $p^{\ell}$ and $\hat{p}^{\ell}$ are the normalized length distributions.

\textbf{JSD-SD.}
For each trajectory, the step distance is
\begin{equation}
s_t(T)=d_{\mathrm{geo}}\!\big((\phi_{t-1},\lambda_{t-1}),(\phi_t,\lambda_t)\big),
~ t=2,\dots,L.
\end{equation}
Pooling all step distances from the real and generated sets yields distributions $p^{\mathrm{step}}$ and $\hat{p}^{\mathrm{step}}$, and we report
\begin{equation}
\mathrm{JSD\text{-}SD}
=
\mathrm{JSD}\!\left(p^{\mathrm{step}},\hat{p}^{\mathrm{step}}\right).
\end{equation}

\subsubsection{Trajectory-level geometric similarity.}
A direct index-wise pairing between real and generated trajectories is inappropriate for generative evaluation, because generated samples are not expected to correspond to real samples one by one. CityTrajBench therefore uses nearest-neighbor matching in a surrogate feature space as a support-based plausibility measure rather than as a reconstruction metric.

Each trajectory is embedded using start location, end location, total length, and a low-resolution polyline representation. In implementation, KD-tree candidate retrieval uses standardized start/end coordinates and total length as summary features, after which candidates are re-ranked by surrogate polyline distance. For each generated trajectory \(\widehat{T}_j\), we retrieve a small candidate set of real trajectories and then select the nearest real trajectory in this surrogate space. Denoting the matched real sample by
\begin{equation}
T_{\pi(j)}=\mathrm{NN}(\widehat{T}_j),
\end{equation}
we compute
\begin{equation}
\mathrm{DTW}_j=\mathrm{DTW}\!\left(T_{\pi(j)},\widehat{T}_j\right),
\end{equation}
\begin{equation}
\mathrm{Frechet}_j=d_F\!\left(T_{\pi(j)},\widehat{T}_j\right).
\end{equation}

This matching strategy is intended to estimate whether a generated trajectory lies near the support of realistic trajectories in the held-out test set, rather than to measure one-to-one reconstruction accuracy or full distributional discrepancy. It should therefore be interpreted jointly with the benchmark's global- and trip-level distribution metrics. For each model, we summarize \(\{\mathrm{DTW}_j\}\) and \(\{\mathrm{Frechet}_j\}\) using the median, 10th percentile (P10), and 90th percentile (P90).

% \paragraph{Conditional mobility consistency.}
\subsubsection{Conditional mobility consistency.}
To evaluate whether generated trajectories preserve origin-dependent destination structure, we partition the dataset bounding box into a shared \(16\times16\) grid and assign each trajectory a start cell \(o\) and an end cell \(d\). For each valid origin cell \(o\) with at least five real trajectories, we construct the conditional destination distribution
\begin{equation}
p(d\mid o)
=
\frac{N_{\mathrm{real}}(o,d)}{\sum_{d'}N_{\mathrm{real}}(o,d')},
\end{equation}
\begin{equation}
\hat{p}(d\mid o)
=
\frac{N_{\mathrm{gen}}(o,d)}{\sum_{d'}N_{\mathrm{gen}}(o,d')}
\end{equation}
where \(N_{\mathrm{real}}(o,d)\) and \(N_{\mathrm{gen}}(o,d)\) denote the numbers of real and generated trajectories whose start and end cells are \((o,d)\), respectively.

For each valid origin cell, we compute the conditional divergence
\begin{equation}
\mathrm{CondDestError}(o)
=
\mathrm{JSD}\!\bigl(p(\cdot\mid o), \hat{p}(\cdot\mid o)\bigr).
\end{equation}
The reported average conditional destination error is
\begin{equation}
\mathrm{AvgCondDestError}
=
\frac{1}{|\mathcal{O}_{\mathrm{valid}}|}
\sum_{o\in\mathcal{O}_{\mathrm{valid}}}
\mathrm{CondDestError}(o),
\end{equation}
where \(\mathcal{O}_{\mathrm{valid}}\) is the set of valid origin cells.

To analyze performance across different origin-frequency regimes, valid origin cells are ranked by their real-data start frequency and divided into three groups: \emph{Head} (top 20\%), \emph{Torso} (middle 30\%), and \emph{Tail} (bottom 50\%).

\textbf{Computational efficiency.}
To complement fidelity-oriented metrics, CityTrajBench also records trainable parameter count, training time, generation latency, throughput, and peak memory footprint. These measurements allow the benchmark to expose fidelity--efficiency trade-offs instead of ranking models only by generation realism.

\textbf{Implementation consistency.}
For a given dataset, all models are evaluated using the same geographic bounds, spatial partition, histogram-construction rule, and nearest-neighbor matching protocol. This benchmark-side consistency is essential for making cross-model differences interpretable.

\section{Datasets, Baselines, and Experimental Protocol}

To enable systematic and reproducible evaluation of city-scale trajectory generation methods, we build CityTrajBench on top of multiple real-world datasets, representative baseline families, and a unified evaluation protocol. This section introduces the datasets, the preprocessing and representation pipeline, the compared baselines, and the fairness constraints used throughout the benchmark.

\subsection{Datasets}

We evaluate CityTrajBench on three real-world trajectory datasets that differ in city context, vehicle type, and mobility regime. As summarized in Table~\ref{tab:dataset_summary}, the benchmark covers two large-scale taxi datasets, i.e., Porto and Chengdu, and one electric-vehicle dataset, i.e., Shanghai. Such diversity enables evaluation under heterogeneous urban settings rather than a single city-specific data distribution.

\begin{table}[h]
\centering
\caption{Summary of the trajectory datasets used in the benchmark.}
\label{tab:dataset_summary}
\setlength{\tabcolsep}{4pt}
\renewcommand{\arraystretch}{1.12}
\begin{tabularx}{\linewidth}{
>{\raggedright\arraybackslash}p{0.14\linewidth}
>{\raggedright\arraybackslash}p{0.10\linewidth}
>{\centering\arraybackslash}p{0.10\linewidth}
>{\centering\arraybackslash}p{0.14\linewidth}
>{\centering\arraybackslash}p{0.10\linewidth}
>{\raggedright\arraybackslash}X}
    \toprule
    Dataset & City & \#Traj & Avg. Trip Distance & Split & Description \\
    \midrule
    Chengdu Taxi & Chengdu & 531517 & 8161.68 m & 70\% /15\% /15\% & Large-scale taxi trips in a dense inland road network \\
    Porto Taxi & Porto & 538495 & 5569.47 m & 70\% /15\% /15\% & Long-term taxi mobility with diverse city-wide route patterns \\
    Shanghai EV & Shanghai & 39615 & 6728.37 m & 70\% /15\% /15\% & EV trips with distinct temporal and mobility regimes \\
    \bottomrule
\end{tabularx}
\end{table}

\textbf{Porto.}
The Porto dataset contains GPS trajectories collected from 442 taxis in Porto, Portugal, from January 2013 to June 2014. It is widely used in trajectory learning research and captures long-term urban taxi mobility over a large spatial region. As a benchmark dataset, it is particularly suitable for evaluating whether generation models can recover large-scale spatial coverage and route diversity.

\textbf{Chengdu.}
The Chengdu dataset contains taxi trajectories collected in Chengdu, China, during October 2016. Following standard cleaning and filtering, it provides a large-scale benchmark with a road topology and travel distribution different from Porto. This dataset is useful for evaluating the robustness of generation models under a different urban structure and traffic organization.

\textbf{Shanghai.}
The Shanghai dataset is derived from electric-vehicle mobility records in Shanghai, China. It contains cleaned trip trajectories from 1,290 EVs over a three-month period, covering different vehicle categories such as battery electric vehicles (BEVs), plug-in hybrid electric vehicles (PHEVs), and range-extended electric vehicles (REEVs). The trajectories are resampled to a uniform 30-second interval. Compared with the taxi datasets, it reflects a different mobility regime and provides a more challenging setting for trajectory generation.

Taken together, these three datasets cover different cities, fleet types, travel behaviors, and road structures, thereby supporting a more comprehensive benchmark for city-scale trajectory generation.

\subsection{Preprocessing and Unified Representation}

All datasets are processed under the CityTrajBench preprocessing pipeline described in Sec.~\ref{subsec:ingest}. In particular, trajectories are split by \texttt{moveid}, normalized to a fixed length of \(L=200\), and mapped to the benchmark coordinate range using dataset-specific geographic bounds. Across all datasets, trajectories are represented under a unified 30-second sampling setting. For datasets with available raw timestamps such as Shanghai, the original records are additionally resampled to this interval; for Porto and Chengdu, the processed benchmark inputs already follow the same 30-second sampling assumption. Trip-level descriptors are then constructed under the same benchmark protocol. In particular, duration- and speed-related features are computed from the unified 30-second representation, while departure-time slots are retained only when valid timestamp information is available; otherwise, the departure-time feature is set to zero.

\begin{table}[t]
\centering
\caption{Baselines included in the benchmark and their methodological categories.}
\label{tab:baseline_summary}
\resizebox{\linewidth}{!}{
\begin{tabular}{llll}
\toprule
Method & Category & Core Idea & Reference \\
\midrule
Markov & Statistical & Transition-based trajectory generation using Markovian state dynamics & \cite{gambs2012next} \\
VAE & VAE-based & Vanilla variational autoencoder baseline for trajectory generation & \cite{kingma2013auto} \\
TrajVAE & VAE-based & Structured variational generation for trajectories & \cite{chen2021trajvae} \\
TrajGAN & GAN-based & Adversarial trajectory synthesis for realistic mobility generation & \cite{liu2018trajgans} \\
DiffTraj & Diffusion-based & Denoising diffusion generation for trajectory sequences & \cite{zhu2023difftraj} \\
DiffRNTraj & Diffusion-based & Diffusion-based trajectory generation with road-network-aware design & \cite{wei2024diff} \\
TrajFlow & Flow-matching-based & Flow-matching-based trajectory generation with scalable continuous transport & \cite{li2026trajflow} \\
\bottomrule
\end{tabular}
}
\end{table}

\subsection{Baselines}

We include seven representative baselines covering multiple trajectory generation paradigms, including classical stochastic modeling, variational autoencoding, adversarial generation, diffusion-based generation, and flow-matching-based generation. Table~\ref{tab:baseline_summary} summarizes their methodological categories and core ideas.

\textit{Markov} \cite{gambs2012next} is a first-order transition baseline and serves as a lightweight statistical reference for coarse-grained mobility generation. \textit{VAE} \cite{kingma2013auto} and \textit{TrajVAE} \cite{chen2021trajvae} represent latent-variable generative approaches, where the latter is more specifically tailored to trajectory generation. \textit{TrajGAN} \cite{liu2018trajgans} represents adversarial trajectory synthesis. \textit{DiffTraj} \cite{zhu2023difftraj} and \textit{DiffRNTraj} \cite{wei2024diff} represent recent diffusion-based methods, with the latter explicitly incorporating road-network structure into the generation process. \textit{TrajFlow} \cite{li2026trajflow} represents a recent flow-matching-based approach that provides an additional comparison point on the realism--efficiency frontier.

In terms of implementation provenance, \textit{Markov}, \textit{VAE}, \textit{TrajVAE}, and \textit{TrajGAN} are implemented by our team within the CityTrajBench framework, while \textit{DiffTraj}, \textit{DiffRNTraj}, and \textit{TrajFlow} are adapted from their official released implementations. This design balances faithful reproduction of public methods with benchmark-level consistency in preprocessing, evaluation, and reporting.

Together, these baselines span heterogeneous modeling assumptions and output spaces, making them suitable for testing whether CityTrajBench supports fair and informative comparison across diverse trajectory generation methods.

\subsection{Evaluation Protocol and Fairness}

All baselines are trained and evaluated under the CityTrajBench protocol. 
In line with Sec.~\ref{sec:method}, the goal of this protocol is not to force identical model architectures, identical input information, or identical optimization behavior across fundamentally different generators. Instead, CityTrajBench standardizes the \emph{benchmark-side} components that determine cross-model comparability: dataset partitioning, preprocessing, representation conversion, sample-count control, evaluation rules, and reporting. 
Table~\ref{tab:fairness_protocol} summarizes the unified settings adopted throughout the benchmark.

\begin{table}[t]
\centering
\caption{Benchmark-side protocol settings adopted by CityTrajBench for comparable evaluation across heterogeneous baselines.}
\label{tab:fairness_protocol}
\begin{tabular}{lp{0.54\linewidth}}
\toprule
Aspect & Unified Setting \\
\midrule
Dataset split & All methods use the same training / validation / test partition on each dataset. \\
Training data & Each model is trained only on the corresponding training split of the target dataset. \\
Generated sample size & For each model and dataset, the number of generated trajectories is matched to the size of the corresponding test split. \\
Evaluation representation & All generated outputs are converted to a common evaluation representation before metric computation. \\
Metric implementation & All methods are evaluated using the same benchmark-side metric implementations and summary rules. \\
Spatial and histogram settings & Geographic bounds, spatial partitions, and histogram-construction rules are fixed per dataset and shared across methods. \\
Trajectory matching & The same nearest-neighbor alignment protocol is used for DTW/Fr\'echet-based trajectory-level comparison. \\
Quantile reporting & Median, 10th percentile, and 90th percentile are reported consistently for all methods. \\
% Failure handling & Invalid or unparsable outputs, if any, are processed using the same benchmark-side filtering rules. \\
Feature policy & Feature availability is standardized at the benchmark level, while feature usage is documented per model. \\
Model selection & Official checkpointing strategy when explicitly available; otherwise benchmark-default validation-based selection; no test-based selection. \\
Reproducibility & Evaluation scripts, preprocessing rules, and parameter settings are shared uniformly across methods. \\
\bottomrule
\end{tabular}
\end{table}

At the dataset level, all methods use the same training, validation, and test split for a given dataset, and each model is trained only on the corresponding training subset. For evaluation, the number of generated trajectories is matched to the size of the held-out test split. This design reduces sample-size-induced bias in distributional metrics and ensures that all methods are compared against the same test distribution under the same sample budget.

At the representation level, CityTrajBench follows the representation-aware unification principle introduced in Sec.~\ref{subsec:ingest}. Because different baselines operate in different output spaces, all generated samples are converted to a common evaluation representation before metric computation. In particular, benchmark-side post-generation conversion follows the same protocol across models, and trajectory-level geometric evaluation uses the same nearest-neighbor matching procedure for all baselines. This consistency is important because, without a shared conversion and matching rule, cross-model differences could be confounded by representation mismatch rather than the generation mechanism itself.

A separate fairness issue concerns \emph{feature access}. As discussed in Sec.~\ref{subsec:features}, CityTrajBench distinguishes between \emph{feature availability} at the benchmark level and \emph{feature usage} at the model level. The benchmark can expose a common pool of trip-level and road-network-level descriptors, but heterogeneous baselines do not necessarily consume the same subset of inputs. We therefore document feature usage explicitly rather than imposing an artificial one-size-fits-all interface. As summarized in Table~\ref{tab:feature_access}, \textit{VAE}, \textit{TrajVAE}, \textit{TrajGAN}, \textit{DiffTraj}, and \textit{TrajFlow} use GPS trajectories together with trip-level descriptors, including departure time, trip distance, trip duration, trajectory length, average step distance, average speed, and start/end region identifiers. \textit{Markov} uses GPS trajectories with start-region information, while \textit{DiffRNTraj} uses GPS trajectories, trip length, and explicit road-network structure. This benchmark-side transparency improves interpretability of the results while preserving fidelity to the original model interfaces.

\begin{table*}[t]
\centering
\caption{Feature-access settings of the compared baselines under CityTrajBench. ``Trip descriptors'' include departure-time slot, trip distance, trip duration, trajectory length, average step distance, average speed, and start/end region identifiers when applicable. For models marked ``Time-of-day = Yes'', the corresponding input dimension is available in the benchmark interface; for datasets without retained real departure timestamps (e.g., Porto and Chengdu), this field is filled with a zero placeholder. ``Road graph'' indicates explicit access to road-network structure.}

\label{tab:feature_access}
\resizebox{\textwidth}{!}{
\begin{tabular}{lcccccc}
\toprule
Method & GPS Sequence & Trip Descriptors & Start/End IDs & Time-of-day & Road Graph & Notes \\
\midrule
Markov     & Yes & No & Yes & No & No & Transition baseline with start-region information \\
VAE        & Yes & Yes & Yes & Yes & No & Self-implemented benchmark baseline \\
TrajVAE    & Yes & Yes & Yes & Yes & No & Self-implemented benchmark baseline \\
TrajGAN    & Yes & Yes & Yes & Yes & No & Self-implemented benchmark baseline \\
DiffTraj   & Yes & Yes & Yes & Yes & No & Adapted from official implementation \\
DiffRNTraj & Yes & Partial & No & No & Yes & Uses trip length and road-network structure \\
TrajFlow   & Yes & Yes & Yes & Yes & No & Adapted from official implementation \\
\bottomrule
\end{tabular}
}
\end{table*}

Model selection is governed by a benchmark policy that balances reproducibility with faithful method reproduction. When an official implementation explicitly specifies a stopping rule or checkpoint-selection strategy, CityTrajBench retains that strategy as part of the reproduced training pipeline. When such details are unavailable or cannot be unambiguously recovered from the original paper or released code, CityTrajBench applies a benchmark-default validation-based model-selection rule. Accordingly, \textit{TrajFlow} follows the checkpoint-selection behavior implemented in its official released code, while the other deep baselines use the benchmark-default validation policy because equally explicit checkpoint-selection rules are not available for them. No method uses the test split for early stopping, checkpoint selection, or hyperparameter tuning.

To improve robustness against accidental training variance, all reported results are averaged over three random seeds, and all mean$\pm$std values are computed from these repeated runs. For a given dataset, all baselines are evaluated with the same preprocessing rules, post-generation conversion procedure, metric implementations, and summary statistics. In particular, geographic bounds, spatial partitions, histogram-construction rules, and nearest-neighbor trajectory-matching settings are fixed at the benchmark level for each dataset and shared across all compared methods. In this way, CityTrajBench reduces implementation-induced variability, avoids test leakage, and makes cross-model differences more interpretable under a standardized benchmark-side protocol, while still documenting residual model-specific differences in feature access, output space, and training behavior.

Concretely, the current CityTrajBench configuration uses trajectory-level splitting by \texttt{moveid} with a 70\%/15\%/15\% train/validation/test ratio and split seed 42; fixed-length trajectory normalization to \(L=200\) by linear interpolation or prefix truncation; coordinate normalization to \([-1,1]\) using dataset-specific geographic bounds; a \(64\times64\) grid for global spatial metrics; a \(16\times16\) grid for conditional destination evaluation; and three random seeds (42, 52, and 62) for repeated runs. A full configuration summary is provided in Appendix~\ref{app:Configuration_summary}.

\section{Experiments}

\subsection{Evaluation Metrics}

We evaluate trajectory generation quality from five complementary perspectives: global spatial realism, trip-level statistical fidelity, trajectory-level geometric similarity, conditional mobility consistency, and computational efficiency.

At the global level, we report \textit{Density Error} and \textit{Pattern Score}, which measure how well a model recovers the city-wide spatial density distribution and the dominant hotspot structure, respectively. At the trip level, we report \textit{Trip Error}, \textit{Length Error}, and \textit{JSD-SD}, which quantify the fidelity of origin--destination distributions, trip-length distributions, and local step-distance statistics. At the trajectory level, we report nearest-neighbor-based DTW and Fr\'echet distances, summarized by the median, 10th percentile (P10), and 90th percentile (P90), so that we can assess not only the typical geometric quality of generated trajectories but also their lower-tail and upper-tail behavior. Unless otherwise stated, lower values indicate better performance for all metrics except \textit{Pattern Score}, where higher values are better.

In addition, we further evaluate \textit{conditional mobility consistency} through origin-conditioned destination divergence and corresponding visualizations. Finally, since practical synthetic trajectory generation must also consider deployability, we compare parameter count, training cost, generation latency, throughput, and memory footprint.
Formal metric definitions and implementation-level configuration details are given in Sec.~\ref{subsec:metrics} and Appendix~\ref{app:Configuration_summary}.

% ===================== Porto =====================
\begin{table*}[t]
\centering
\caption{Results on Porto}
\label{tab:porto}
\resizebox{\textwidth}{!}{
\begin{tabular}{lcc|ccc|cccccc}
\toprule
\multirow{2}{*}{Model}
& \multicolumn{2}{c|}{Global Level}
& \multicolumn{3}{c|}{Trip Level}
& \multicolumn{6}{c}{Traj Level} \\
\cline{2-12}
& Density Error $\downarrow$ & Pattern Score $\uparrow$
& Trip Error $\downarrow$ & Length Error $\downarrow$ & JSD-SD $\downarrow$
& DTW (Med.) $\downarrow$ & DTW (P10) $\downarrow$ & DTW (P90) $\downarrow$
& Fr\'echet (Med.) $\downarrow$ & Fr\'echet (P10) $\downarrow$ & Fr\'echet (P90) $\downarrow$ \\
\midrule
Markov
& $0.0252 \pm 0.0001$
& $0.8134 \pm 0.0014$
& $\mathbf{0.0658} \pm 0.0008$
& $0.1723 \pm 0.0011$
& $\mathbf{0.000001} \pm 0.000000$
& $92.4712 \pm 0.0310$
& $50.0876 \pm 0.0706$
& $143.4532 \pm 0.2788$
& $0.9174 \pm 0.0018$
& $0.6129 \pm 0.0014$
& $1.3279 \pm 0.0017$ \\
VAE
& $0.1527 \pm 0.0007$
& $0.3871 \pm 0.0093$
& $0.2603 \pm 0.0055$
& $\underline{0.0129} \pm 0.0037$
& $0.001784 \pm 0.000185$
& $50.7349 \pm 1.0318$
& $29.3256 \pm 0.6819$
& $\underline{86.6801} \pm 1.5097$
& $0.5137 \pm 0.0120$
& $0.3177 \pm 0.0063$
& $\underline{0.8482} \pm 0.0170$ \\
TrajVAE
& $0.1725 \pm 0.0029$
& $0.3708 \pm 0.0125$
& $0.3239 \pm 0.0084$
& $0.0790 \pm 0.0152$
& $0.010608 \pm 0.000401$
& $79.9569 \pm 0.8341$
& $41.9737 \pm 0.4429$
& $145.8584 \pm 1.2039$
& $0.7909 \pm 0.0170$
& $0.4319 \pm 0.0065$
& $1.3258 \pm 0.0140$ \\
TrajGAN
& $0.1472 \pm 0.0392$
& $0.4466 \pm 0.0767$
& $0.2611 \pm 0.0467$
& $0.2353 \pm 0.1718$
& $0.011652 \pm 0.010679$
& $98.1752 \pm 34.7511$
& $57.5962 \pm 17.8912$
& $173.6251 \pm 69.9077$
& $0.9766 \pm 0.3304$
& $0.6149 \pm 0.1989$
& $1.6280 \pm 0.6565$ \\
DiffTraj
& $0.0237 \pm 0.0011$
& $\underline{0.8509} \pm 0.0171$
& $0.1172 \pm 0.0086$
& $\mathbf{0.0008} \pm 0.0014$
& $\underline{0.000112} \pm 0.000186$
& $\mathbf{16.0542} \pm 0.7365$
& $\underline{7.1404} \pm 0.3623$
& $\mathbf{48.3437} \pm 1.6281$
& $\mathbf{0.2892} \pm 0.0116$
& $\mathbf{0.1301} \pm 0.0059$
& $\mathbf{0.6809} \pm 0.0194$ \\
DiffRNTraj
& $\underline{0.0204} \pm 0.0002$
& $0.8191 \pm 0.0088$
& $0.0805 \pm 0.0012$
& $0.0228 \pm 0.0005$
& $0.002314 \pm 0.000104$
& $53.6805 \pm 0.6993$
& $12.2674 \pm 0.3300$
& $169.5574 \pm 2.7157$
& $0.6800 \pm 0.0052$
& $0.2618 \pm 0.0044$
& $1.5959 \pm 0.0228$ \\
TrajFlow
& $\mathbf{0.0148} \pm 0.0009$
& $\mathbf{0.8851} \pm 0.0136$
& $\underline{0.0799} \pm 0.0005$
& $0.0288 \pm 0.0186$
& $0.000736 \pm 0.000455$
& $\underline{22.5946} \pm 1.3099$
& $\mathbf{7.1209} \pm 0.2267$
& $95.2365 \pm 6.9026$
& $\underline{0.4243} \pm 0.0172$
& $\underline{0.1426} \pm 0.0036$
& $1.0849 \pm 0.0553$ \\
\bottomrule
\end{tabular}
}
\end{table*}

% ===================== Chengdu =====================
\begin{table*}[t]
\centering
\caption{Results on Chengdu}
\label{tab:chengdu}
\resizebox{\textwidth}{!}{
\begin{tabular}{lcc|ccc|cccccc}
\toprule
\multirow{2}{*}{Model}
& \multicolumn{2}{c|}{Global Level}
& \multicolumn{3}{c|}{Trip Level}
& \multicolumn{6}{c}{Traj Level} \\
\cline{2-12}
& Density Error $\downarrow$ & Pattern Score $\uparrow$
& Trip Error $\downarrow$ & Length Error $\downarrow$ & JSD-SD $\downarrow$
& DTW (Med.) $\downarrow$ & DTW (P10) $\downarrow$ & DTW (P90) $\downarrow$
& Fr\'echet (Med.) $\downarrow$ & Fr\'echet (P10) $\downarrow$ & Fr\'echet (P90) $\downarrow$ \\
\midrule
Markov
& $0.0885 \pm 0.0001$
& $0.7221 \pm 0.0014$
& $0.1126 \pm 0.0002$
& $0.0262 \pm 0.0014$
& $\mathbf{0.000043} \pm 0.000003$
& $156.5843 \pm 0.4329$
& $90.7829 \pm 0.1100$
& $292.4632 \pm 1.1693$
& $1.6027 \pm 0.0028$
& $1.0539 \pm 0.0027$
& $2.5847 \pm 0.0080$ \\
VAE
& $0.1687 \pm 0.0007$
& $0.4800 \pm 0.0056$
& $0.2626 \pm 0.0035$
& $\underline{0.0198} \pm 0.0102$
& $0.008206 \pm 0.000842$
& $55.3763 \pm 1.5477$
& $24.8629 \pm 0.3456$
& $108.5312 \pm 1.8550$
& $0.6154 \pm 0.0206$
& $0.3321 \pm 0.0090$
& $\underline{1.0808} \pm 0.0156$ \\
TrajVAE
& $0.1636 \pm 0.0088$
& $0.5094 \pm 0.0185$
& $0.3666 \pm 0.0066$
& $0.0832 \pm 0.0088$
& $0.016209 \pm 0.001987$
& $86.6598 \pm 2.8439$
& $38.9554 \pm 1.6736$
& $172.2181 \pm 4.4269$
& $0.9749 \pm 0.0194$
& $0.5363 \pm 0.0125$
& $1.6489 \pm 0.0074$ \\
TrajGAN
& $0.1435 \pm 0.0536$
& $0.5265 \pm 0.1497$
& $0.3177 \pm 0.0663$
& $0.1626 \pm 0.0731$
& $0.022803 \pm 0.004870$
& $73.7766 \pm 23.9946$
& $33.2222 \pm 16.8419$
& $161.5605 \pm 27.4732$
& $0.8968 \pm 0.1908$
& $0.4802 \pm 0.1728$
& $1.6872 \pm 0.1834$ \\
DiffTraj
& $\underline{0.0204} \pm 0.0010$
& $\underline{0.8875} \pm 0.0042$
& $0.0932 \pm 0.0117$
& $\mathbf{0.0021} \pm 0.0002$
& $\underline{0.001521} \pm 0.000154$
& $\mathbf{14.4064} \pm 0.3730$
& $\mathbf{5.5332} \pm 0.1855$
& $\mathbf{50.4729} \pm 1.5272$
& $\mathbf{0.3021} \pm 0.0033$
& $\mathbf{0.0969} \pm 0.0021$
& $\mathbf{0.8111} \pm 0.0128$ \\
DiffRNTraj
& $0.0222 \pm 0.0008$
& $0.8354 \pm 0.0099$
& $\underline{0.0808} \pm 0.0007$
& $0.0307 \pm 0.0023$
& $0.003124 \pm 0.000193$
& $44.2145 \pm 0.9063$
& $7.0357 \pm 0.1500$
& $140.9712 \pm 1.5240$
& $0.7600 \pm 0.0070$
& $0.1743 \pm 0.0052$
& $1.6055 \pm 0.0107$ \\
TrajFlow
& $\mathbf{0.0141} \pm 0.0011$
& $\mathbf{0.9177} \pm 0.0116$
& $\mathbf{0.0750} \pm 0.0078$
& $0.0397 \pm 0.0101$
& $0.005237 \pm 0.000943$
& $\underline{18.8813} \pm 1.8762$
& $\underline{5.8942} \pm 0.1638$
& $\underline{84.8509} \pm 7.5672$
& $\underline{0.4152} \pm 0.0352$
& $\underline{0.1094} \pm 0.0057$
& $1.2050 \pm 0.0649$ \\
\bottomrule
\end{tabular}
}
\end{table*}

% ============ Shanghai ============
\begin{table*}[t]
\centering
\caption{Results on Shanghai}
\label{tab:shanghai}
\resizebox{\textwidth}{!}{
\begin{tabular}{lcc|ccc|cccccc}
\toprule
\multirow{2}{*}{Model}
& \multicolumn{2}{c|}{Global Level}
& \multicolumn{3}{c|}{Trip Level}
& \multicolumn{6}{c}{Traj Level} \\
\cline{2-12}
& Density Error $\downarrow$ & Pattern Score $\uparrow$
& Trip Error $\downarrow$ & Length Error $\downarrow$ & JSD-SD $\downarrow$
& DTW (Med.) $\downarrow$ & DTW (P10) $\downarrow$ & DTW (P90) $\downarrow$
& Fr\'echet (Med.) $\downarrow$ & Fr\'echet (P10) $\downarrow$ & Fr\'echet (P90) $\downarrow$ \\
\midrule
Markov
& $\underline{0.0694} \pm 0.0011$
& $0.7042 \pm 0.0065$
& $\mathbf{0.1508} \pm 0.0038$
& $0.1534 \pm 0.0004$
& $\mathbf{0.000038} \pm 0.000006$
& $66.7628 \pm 0.4947$
& $20.6044 \pm 0.8662$
& $\mathbf{138.7191} \pm 0.8883$
& $0.7135 \pm 0.0029$
& $0.2925 \pm 0.0056$
& $\mathbf{1.3919} \pm 0.0184$ \\
VAE
& $0.1912 \pm 0.0092$
& $0.2877 \pm 0.0334$
& $0.4044 \pm 0.0107$
& $0.2408 \pm 0.0455$
& $0.000786 \pm 0.000138$
& $111.9844 \pm 15.5943$
& $50.5049 \pm 12.3403$
& $193.4829 \pm 19.0579$
& $1.0809 \pm 0.1611$
& $0.5225 \pm 0.1113$
& $1.9079 \pm 0.1661$ \\
TrajVAE
& $0.2097 \pm 0.0043$
& $0.3040 \pm 0.0099$
& $0.5047 \pm 0.0139$
& $0.1196 \pm 0.0124$
& $0.000164 \pm 0.000065$
& $85.1678 \pm 2.5113$
& $40.4814 \pm 0.9462$
& $\underline{151.8149} \pm 3.8028$
& $0.9047 \pm 0.0272$
& $0.5012 \pm 0.0243$
& $\underline{1.4928} \pm 0.0301$ \\
TrajGAN
& $0.2289 \pm 0.0769$
& $0.2698 \pm 0.1813$
& $0.4714 \pm 0.1269$
& $0.3291 \pm 0.3119$
& $0.044282 \pm 0.076102$
& $203.0996 \pm 224.5991$
& $127.5870 \pm 165.9713$
& $302.6424 \pm 251.9127$
& $2.0038 \pm 2.1734$
& $1.3529 \pm 1.6633$
& $2.7665 \pm 2.0781$ \\
DiffTraj
& $0.1584 \pm 0.0273$
& $0.4849 \pm 0.0882$
& $0.3367 \pm 0.0361$
& $\mathbf{0.0135} \pm 0.0116$
& $\underline{0.000042} \pm 0.000018$
& $\underline{58.4628} \pm 12.0871$
& $\underline{19.4879} \pm 8.2192$
& $170.6667 \pm 11.0149$
& $\mathbf{0.5639} \pm 0.0814$
& $\underline{0.2267} \pm 0.0654$
& $1.7169 \pm 0.0647$ \\
DiffRNTraj
& $\mathbf{0.0594} \pm 0.0020$
& $\mathbf{0.7751} \pm 0.0049$
& $0.2346 \pm 0.0026$
& $0.0720 \pm 0.0005$
& $0.006943 \pm 0.000364$
& $100.1708 \pm 2.2021$
& $24.9288 \pm 0.7315$
& $281.7933 \pm 1.2137$
& $1.1115 \pm 0.0107$
& $0.3128 \pm 0.0040$
& $2.7834 \pm 0.0291$ \\
TrajFlow
& $0.0782 \pm 0.0032$
& $\underline{0.7148} \pm 0.0166$
& $\underline{0.2130} \pm 0.0025$
& $\underline{0.0630} \pm 0.0060$
& $0.000271 \pm 0.000039$
& $\mathbf{51.0149} \pm 3.0897$
& $\mathbf{7.2631} \pm 0.1509$
& $172.4402 \pm 8.9677$
& $\underline{0.5971} \pm 0.0255$
& $\mathbf{0.1220} \pm 0.0030$
& $1.7590 \pm 0.0878$ \\
\bottomrule
\end{tabular}
}
\end{table*}

\subsection{Quantitative Results}
Tables~\ref{tab:porto}--\ref{tab:shanghai} summarize the benchmark results on Porto, Chengdu, and Shanghai, where  the best and second-best values in each column are highlighted in bold and underline, respectively.
Several consistent patterns emerge.
First, the metrics are clearly complementary rather than redundant. Models that perform strongly on city-scale density realism do not necessarily achieve the best trajectory-level geometric similarity, and models that preserve endpoint or local step statistics well do not automatically recover realistic full-route shapes. This validates the design of CityTrajBench: urban trajectory generation quality cannot be characterized by a single metric family. Because these metric families capture different and sometimes competing aspects of generation quality, we do not collapse them into a single aggregate score. 

Second, the strongest results are delivered by continuous generative models, but with distinct specializations. 
\textit{DiffTraj} is the most consistently strong model on trajectory-level fidelity across datasets, indicating superior recovery of fine-grained route geometry under the current benchmark protocol. 
\textit{DiffRNTraj} is particularly competitive on structure-sensitive global metrics, suggesting that explicit road-network constraints help preserve city-scale spatial organization and major travel corridors. 
\textit{TrajFlow} also emerges as a highly competitive model, often reaching the best or second-best results on global metrics while remaining strong at the trajectory level.

Third, simpler baselines remain informative. In particular, \textit{Markov} is consistently competitive on \textit{Trip Error} and \textit{JSD-SD}, which implies that endpoint distributions and local movement increments are still substantially governed by low-order mobility regularities. However, its weakness on DTW and Fr\'echet shows that matching coarse statistics is not sufficient for recovering realistic route-level behavior.

Finally, result variance across random seeds provides additional insight. \textit{TrajGAN} exhibits noticeably larger fluctuations than the other methods, especially on Shanghai, indicating weaker optimization stability and less reliable reproducibility. By contrast, the diffusion- and flow-based models are generally more stable across repeated runs.

\subsubsection{Results on Porto}

Table~\ref{tab:porto} shows that Porto already exposes a clear separation between three notions of generation quality: coarse statistical realism, structure-aware spatial realism, and trajectory-level geometric fidelity.

At the global and trip levels, the best results are distributed across different models rather than dominated by a single one. \textit{TrajFlow} achieves the best \textit{Density Error} (0.0148) and \textit{Pattern Score} (0.8851), indicating the strongest recovery of city-wide density structure and hotspot layout. \textit{Markov} obtains the best \textit{Trip Error} (0.0658) and \textit{JSD-SD} ($1\times10^{-6}$), showing that a simple transition-based model can still reproduce endpoint distributions and local movement scales very well. \textit{DiffTraj} attains the best \textit{Length Error} (0.0008), suggesting especially strong fidelity in trip-scale travel distance.

At the trajectory level, however, the ranking changes sharply. \textit{DiffTraj} is the clear best model on all six DTW and Fr\'echet statistics, with a substantial margin over the remaining baselines. This indicates that its denoising process is particularly effective at recovering individual route geometry rather than merely matching aggregate distributions. \textit{TrajFlow} forms the second strongest group on Porto, with competitive \textit{DTW (Med.)}, \textit{DTW (P10)}, \textit{Fr\'echet (Med.)}, and \textit{Fr\'echet (P10)}, while \textit{VAE} is better than \textit{TrajVAE} and \textit{TrajGAN} but still clearly behind \textit{DiffTraj}.

An important implication from Porto is that strong macro-level realism does not automatically imply strong sample-level fidelity. For example, \textit{TrajFlow} and \textit{DiffRNTraj} recover city-scale structure well, yet only \textit{DiffTraj} simultaneously achieves very low trajectory-level distances. This suggests that hotspot recovery and road conformity alone are insufficient to guarantee realistic route-shape generation.

\subsubsection{Results on Chengdu}

Table~\ref{tab:chengdu} reveals a pattern highly consistent with Porto, which suggests that the benchmark is capturing stable model characteristics rather than dataset-specific noise.

At the global and trip levels, \textit{TrajFlow} is the strongest model overall on Chengdu. It achieves the best \textit{Density Error} (0.0141), \textit{Pattern Score} (0.9177), and \textit{Trip Error} (0.0750), indicating excellent recovery of spatial density, hotspot structure, and endpoint distributions. \textit{DiffTraj} again performs best on \textit{Length Error} (0.0021), showing strong control over trip-scale travel distances, while \textit{Markov} retains the best \textit{JSD-SD} ($4.3\times10^{-5}$), reinforcing the observation that local step statistics are comparatively easy to match with low-order transition mechanisms.

At the trajectory level, \textit{DiffTraj} again dominates all DTW and Fr\'echet statistics. Importantly, this advantage is not limited to the median. Its \textit{DTW (P90)} and \textit{Fr\'echet (P90)} are also substantially lower than those of all other methods, which means that \textit{DiffTraj} not only improves typical-case trajectory quality but also reduces the tail risk of generating severely unrealistic routes. In contrast, \textit{DiffRNTraj} remains strong on global realism but is much weaker on route-level geometry, while \textit{TrajFlow} offers a more balanced profile but still does not fully match \textit{DiffTraj} on fine-grained trajectory similarity.

Compared with Porto, Chengdu makes the specialization pattern even clearer. \textit{TrajFlow} and \textit{DiffRNTraj} are especially strong at reproducing city-scale organization, whereas \textit{DiffTraj} is especially strong at reproducing individual route shapes. This distinction is precisely the kind of trade-off that would be obscured by a benchmark built around only one metric family.

\subsubsection{Results on Shanghai}

Table~\ref{tab:shanghai} reports the results on Shanghai, which is the smallest dataset in our benchmark and also the one with the most mixed ranking pattern. It therefore serves as a stress test for robustness under a different mobility regime and more challenging data conditions.

At the global and trip levels, \textit{DiffRNTraj} becomes the strongest structure-aware model, achieving the best \textit{Density Error} (0.0594) and \textit{Pattern Score} (0.7751). \textit{TrajFlow} remains highly competitive, ranking second on \textit{Pattern Score}, \textit{Trip Error}, and \textit{Length Error}. Meanwhile, \textit{Markov} again obtains the best \textit{Trip Error} (0.1508) and \textit{JSD-SD} ($3.8\times10^{-5}$), which shows that even in this harder setting, low-order transition structure still explains part of the endpoint and local-step behavior.

At the trajectory level, the ranking becomes less uniform than on Porto and Chengdu. \textit{TrajFlow} achieves the best \textit{DTW (Med.)}, \textit{DTW (P10)}, and \textit{Fr\'echet (P10)}, while \textit{DiffTraj} achieves the best \textit{Fr\'echet (Med.)}. By contrast, \textit{Markov} and \textit{TrajVAE} perform relatively well on some upper-tail statistics such as \textit{DTW (P90)} and \textit{Fr\'echet (P90)}. This indicates that under Shanghai, model quality is no longer cleanly ordered by a single ranking and becomes more sensitive to whether one emphasizes average-case geometry, local alignment, or worst-case failures.

This shift in ranking is itself informative. Compared with the two taxi datasets, Shanghai weakens the dominance of any one method and exposes a clearer tension between global realism and trajectory-level fidelity. One plausible benchmark-level  explanation is that the EV-oriented dataset, together with its smaller scale and distinct sampling regime, yields a less regular mobility distribution under the current benchmark representation. Under such conditions, preserving city-scale structure and preserving sample-level geometry may become harder to optimize simultaneously. We emphasize, however, that this interpretation is inferential rather than causal, and should be read as a data-dependent hypothesis consistent with the observed benchmark pattern.

\subsubsection{Cross-Dataset Findings}

Across the three datasets, four benchmark-level findings consistently emerge.

\textbf{(1) Different metric families probe genuinely different capabilities.} Models that perform well on \textit{Density Error} and \textit{Pattern Score} do not necessarily achieve the best DTW/Fr\'echet statistics, and vice versa. This confirms that trajectory generation quality is inherently multi-dimensional.

\textbf{(2) Diffusion-style and flow-style models currently define the strongest frontier, but with different strengths.} \textit{DiffTraj} is the most consistently strong model for route-level geometric fidelity. \textit{DiffRNTraj} is especially strong on structure-sensitive macro realism. \textit{TrajFlow} is competitive in both dimensions and is often strongest on global-level metrics. Together, these methods occupy different points on the realism trade-off frontier rather than forming a single strict ranking.

\textbf{(3) Explicit structural priors help macro realism more than micro geometry.} The strong performance of \textit{DiffRNTraj} on density and pattern metrics suggests that road-network constraints compress the feasible generation space in a useful way. However, this structural advantage does not automatically translate into the best route-level similarity. In other words, topological plausibility helps recover city-scale organization, but may also reduce flexibility in matching fine-grained geometric details.

\begin{figure*}[htbp]
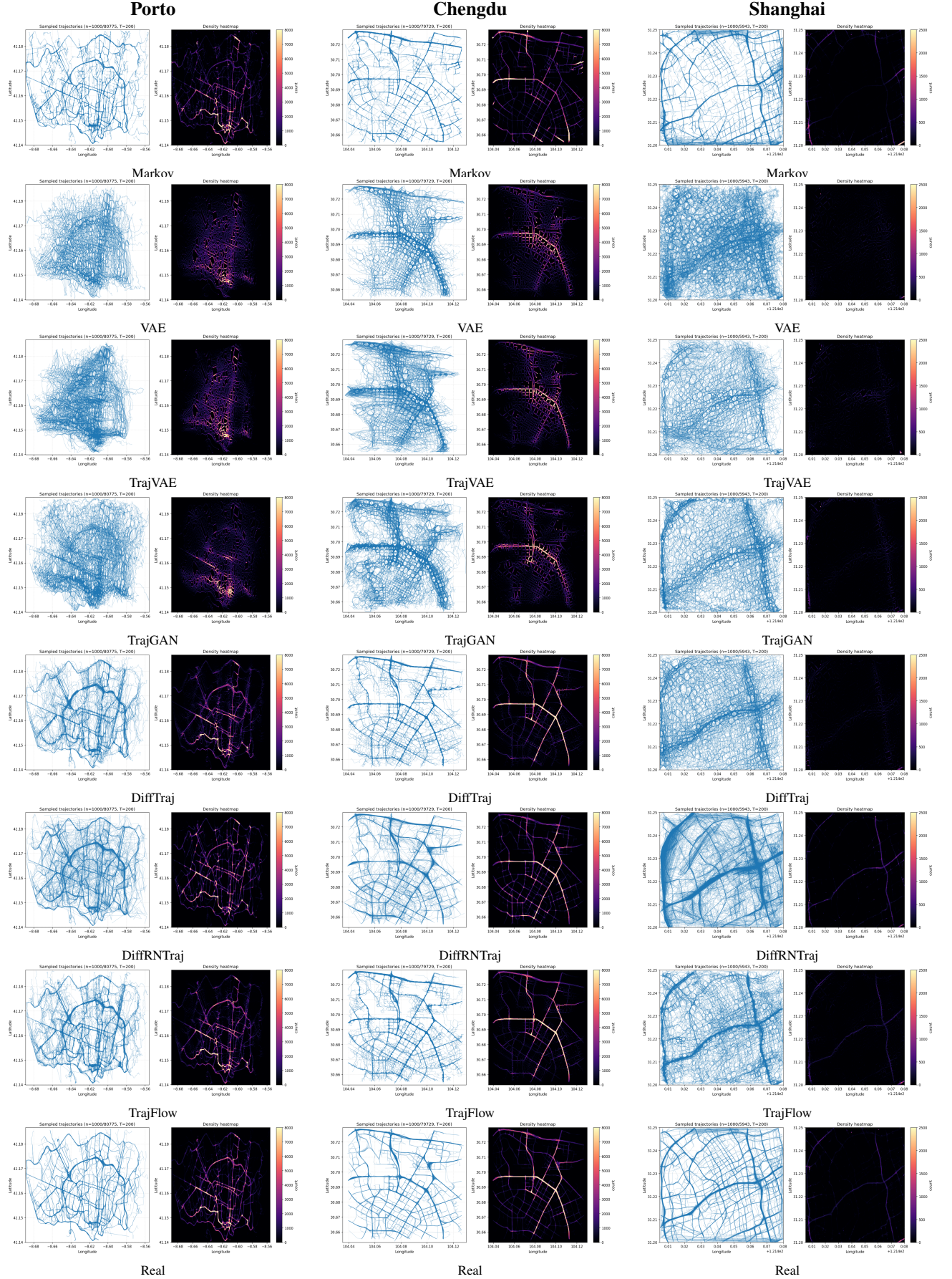

\centering
\setlength{\tabcolsep}{6pt}
\renewcommand{\arraystretch}{1.1}
\begin{tabular}{@{}c c c@{}}
\textbf{Porto} & \textbf{Chengdu} & \textbf{Shanghai} \\

\imgcell{Porto}{Markov} & \imgcell{Chengdu}{Markov} & \imgcell{Shanghai_Small_Enhanced}{Markov} \\
\imgcell{Porto}{VAE} & \imgcell{Chengdu}{VAE} & \imgcell{Shanghai_Small_Enhanced}{VAE} \\
\imgcell{Porto}{TrajVAE} & \imgcell{Chengdu}{TrajVAE} & \imgcell{Shanghai_Small_Enhanced}{TrajVAE} \\
\imgcell{Porto}{TrajGAN} & \imgcell{Chengdu}{TrajGAN} & \imgcell{Shanghai_Small_Enhanced}{TrajGAN} \\
\imgcell{Porto}{DiffTraj} & \imgcell{Chengdu}{DiffTraj} & \imgcell{Shanghai_Small_Enhanced}{DiffTraj} \\
\imgcell{Porto}{DiffRNTraj} & \imgcell{Chengdu}{DiffRNTraj} & \imgcell{Shanghai_Small_Enhanced}{DiffRNTraj} \\
\imgcell{Porto}{TrajFlow} & \imgcell{Chengdu}{TrajFlow} & \imgcell{Shanghai_Small_Enhanced}{TrajFlow} \\
\imgcell{Porto}{Real} & \imgcell{Chengdu}{Real} & \imgcell{Shanghai_Small_Enhanced}{Real} \\
\end{tabular}
\caption{Qualitative overview of generated and real trajectories across Porto, Chengdu, and Shanghai. 
For each model and dataset, each panel contains representative trajectory samples and the corresponding spatial density map, enabling side-by-side inspection of route geometry and global occupancy structure.}

\label{fig:visualization}
\end{figure*}

\textbf{(4) Low-order mobility regularity remains surprisingly strong.} The competitiveness of \textit{Markov} on \textit{Trip Error} and \textit{JSD-SD} across all three datasets indicates that endpoint and local-step distributions are easier to capture than full-route geometry. This also explains why single-metric evaluation can be misleading: a method may appear strong if only OD realism is measured, while still generating unrealistic full trajectories.

\subsection{Qualitative Analysis}

Figure~\ref{fig:visualization} visualizes the real and generated trajectories on the three datasets. The qualitative patterns are broadly consistent with the quantitative findings, but they also reveal several failure modes that are less obvious from the tables alone.

First, \textit{DiffRNTraj} better recovers the major urban corridors and high-density spatial skeletons than the VAE-, GAN-, and Markov-based baselines. Its generated trajectories align more closely with dominant road structures and hotspot regions, which is consistent with its strong performance on \textit{Density Error} and \textit{Pattern Score}. Visually, it tends to produce more network-conforming trajectory sets, reflecting the benefit of explicit road-network awareness.

Second, \textit{DiffTraj} produces trajectories that are more plausible at the individual-route level. Compared with the more structure-constrained or coarse statistical baselines, its samples often exhibit better local continuity and route-shape realism. This observation aligns with its clear advantage on DTW and Fr\'echet metrics, and suggests that \textit{DiffTraj} is particularly effective at modeling the geometry of complete paths rather than only the aggregate occupancy pattern.

Third, \textit{Markov} captures several major travel directions but tends to over-concentrate movement into coarse and repetitive transition patterns. This helps explain why it remains competitive on \textit{Trip Error} yet weak on trajectory-level distances. The VAE-based and GAN-based baselines can produce locally plausible trajectories in some regions, but they generally underperform in overall spatial coverage and corridor recovery. Their samples often appear either overly diffuse or insufficiently aligned with the dominant urban structure, especially on Shanghai.

Overall, the visual evidence reinforces the quantitative conclusion that different models specialize in different notions of realism: \textit{DiffRNTraj} is stronger at recovering the city-scale spatial skeleton, whereas \textit{DiffTraj} is stronger at recovering the geometry of individual trips.

\subsection{Conditional Mobility Consistency}
\label{sec:Conditional_Mobility_Consistency}
\begin{figure}[h]
\centering
\includegraphics[width=\linewidth]{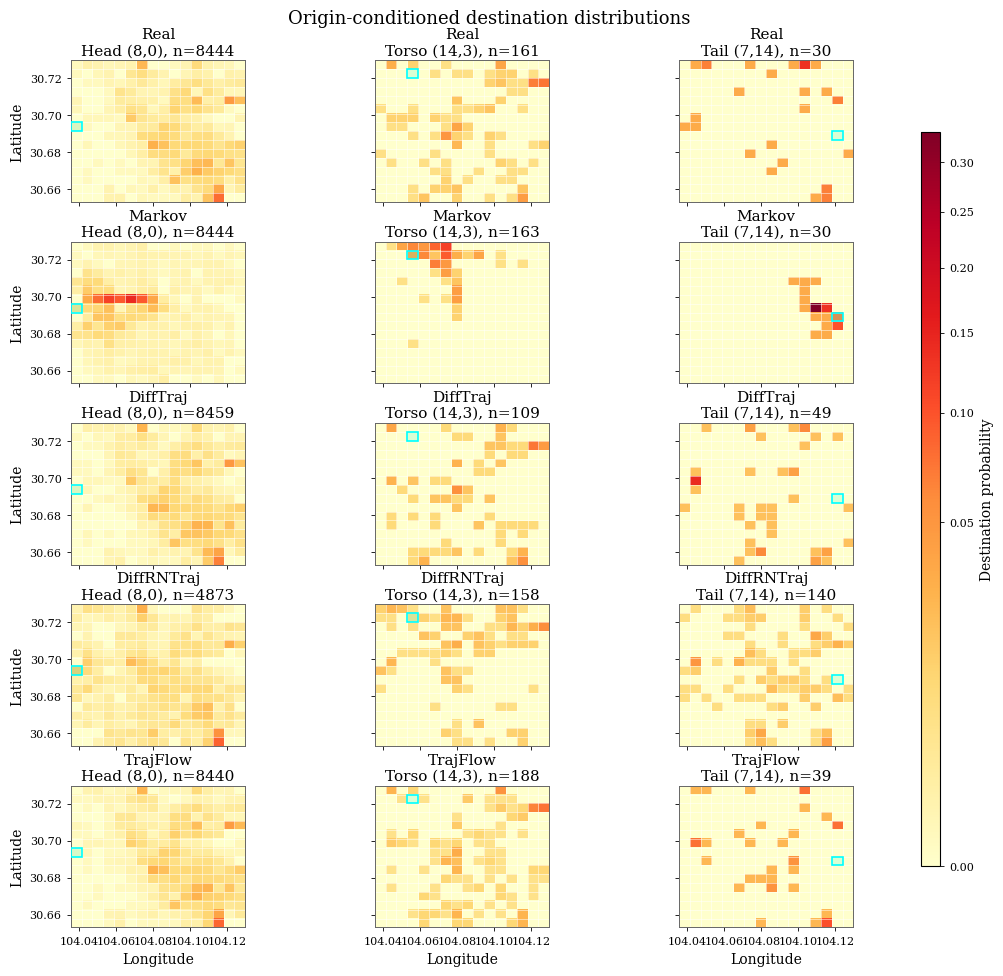}
\caption{Origin-conditioned destination distributions on Chengdu for representative head, torso, and tail origin cells. Each panel compares the real and generated destination distributions under the same origin condition.}
\label{fig:origin_conditioned_destination}
\end{figure}

\begin{figure}[h]
\centering
\includegraphics[width=\linewidth]{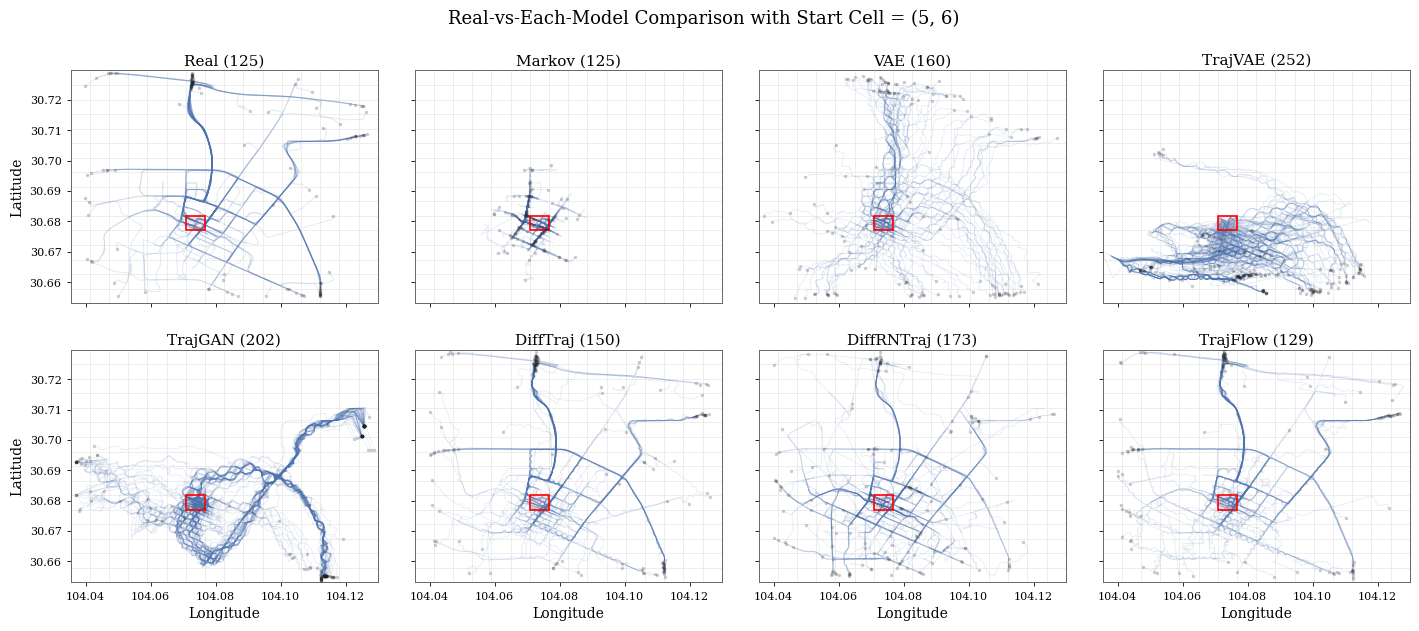}
\caption{Origin-conditioned trajectory examples on Chengdu. Under the same origin condition, each panel compares real and generated trajectory samples to illustrate downstream spatial spread and directional diversity across models.}
\label{fig:Origin-conditioned_Trajectory}
\end{figure}

We next evaluate whether the models preserve conditional dependence between origins and downstream mobility patterns. This evaluation should be interpreted under the current benchmark feature-access and model-adaptation setting. In particular, compared baselines do not use identical conditioning interfaces, so the reported conditional results reflect both generation quality and the extent to which each reproduced model can exploit origin-related information under CityTrajBench.
We report the main conditional destination evaluation on Chengdu, which provides a clear and stable setting for head/torso/tail analysis and representative visualization under the current protocol. Additional conditional results on Porto and Shanghai are included in Appendix~\ref{app:Additional Conditional Destination Results on Porto and Shanghai} for completeness.
Table~\ref{tab:cond_dest} reports origin-conditioned destination divergence, and Figures~\ref{fig:origin_conditioned_destination} and~\ref{fig:Origin-conditioned_Trajectory} provide representative visual examples. In contrast to the earlier benchmark version, the current conditional evaluation also includes \textit{TrajFlow}, whose reproduced implementation now supports the corresponding benchmark protocol.

\begin{table}[h]
% \caption{Origin-conditioned destination divergence}
\caption{Origin-conditioned destination divergence on Chengdu.}
\label{tab:cond_dest}
\centering
\resizebox{\linewidth}{!}{%
\begin{tabular}{lccccc}
\toprule
     Model &  AvgCondDestError &   Head &  Torso &   Tail &  NumOrigins \\
\midrule
    Markov &            0.6501 & 0.6406 & 0.6496 & 0.6543 &         246 \\
       VAE &            0.3837 & 0.2668 & 0.3508 & 0.4507 &         246 \\
   TrajVAE &            0.5163 & 0.4497 & 0.4813 & 0.5642 &         246 \\
   TrajGAN &            0.5505 & 0.5058 & 0.5366 & 0.5768 &         246 \\
  DiffTraj &            \textbf{0.1532} & \underline{0.0563} & \textbf{0.1037} & \textbf{0.2220} &         246 \\
DiffRNTraj &            0.3740 & 0.2261 & 0.3352 & 0.4572 &         246 \\
  TrajFlow &            \underline{0.1585} & \textbf{0.0530} & \underline{0.1096} & \underline{0.2304} &         246 \\
\bottomrule
\end{tabular}
}
\end{table}

A first observation is that \textit{DiffTraj} and \textit{TrajFlow} clearly form the strongest pair under origin-conditioned evaluation. \textit{DiffTraj} achieves the lowest overall \textit{AvgCondDestError} (0.1532), followed very closely by \textit{TrajFlow} (0.1585), while the remaining methods are separated by a large margin. This finding is important because conditional realism is more demanding than unconditional density matching: a model must not only place probability mass in plausible regions globally, but also place it in the correct regions given a specific starting context.

At the regime level, the comparison between \textit{DiffTraj} and \textit{TrajFlow} is even more informative. On \textit{Head} origins, \textit{TrajFlow} achieves the best result (0.0530), slightly outperforming \textit{DiffTraj} (0.0563). On \textit{Torso} and \textit{Tail} origins, however, \textit{DiffTraj} regains the lead with the lowest divergence values, i.e., 0.1037 on \textit{Torso} and 0.2220 on \textit{Tail}, compared with 0.1096 and 0.2304 for \textit{TrajFlow}, respectively. This suggests that \textit{TrajFlow} is highly effective for high-frequency origin regimes, whereas \textit{DiffTraj} remains slightly more robust when the conditional destination structure becomes sparser and more difficult to recover.

The visual comparisons in Fig.~\ref{fig:origin_conditioned_destination} support the same conclusion. For frequent origins (\textit{Head}), the real data exhibit a relatively broad and multi-modal destination distribution. Both \textit{DiffTraj} and \textit{TrajFlow} recover this structure much better than the remaining baselines, with concentrated hotspots appearing in the correct general regions. In contrast, \textit{Markov} still collapses much of the conditional mass into a narrower local area, while \textit{DiffRNTraj} produces a more diffuse pattern with weaker hotspot sharpness.

For medium-frequency origins (\textit{Torso}), the real destination pattern becomes more scattered and sparse. Here, \textit{DiffTraj} provides the closest visual match to the real support, with \textit{TrajFlow} remaining highly competitive and substantially better than the other baselines. \textit{Markov}, \textit{VAE}, and \textit{TrajVAE} still exhibit stronger aggregation bias, while \textit{TrajGAN} appears less stable and less well calibrated. \textit{DiffRNTraj} captures part of the dispersion, but its destination mass is still more diffuse than that of \textit{DiffTraj} and \textit{TrajFlow}.

For rare origins (\textit{Tail}), the task becomes particularly challenging because the real destination distribution is highly sparse and long-tailed. In this regime, \textit{DiffTraj} again provides the best match, and \textit{TrajFlow} remains the second strongest model. Both are clearly better than the remaining baselines at preserving sparse long-range destination support. By contrast, \textit{Markov} and the VAE/GAN baselines allocate too much mass to nearby cells or fail to recover the weak but meaningful long-range destination structure. \textit{DiffRNTraj} performs better than these weaker baselines, but still appears more over-diffused than the two strongest continuous generative models.

Figure~\ref{fig:Origin-conditioned_Trajectory} further illustrates the same pattern at the trajectory level. Under the same origin condition, \textit{DiffTraj} and \textit{TrajFlow} generate trajectories whose downstream spatial spread and directional diversity better resemble the real samples, whereas \textit{Markov} and several weaker baselines show either excessive concentration or insufficiently structured dispersion. This indicates that stronger conditional destination calibration is accompanied by stronger origin-conditioned trajectory realism.

Overall, the updated conditional benchmark refines the earlier global findings. While \textit{DiffRNTraj} remains highly competitive on structure-sensitive macro metrics, \textit{DiffTraj} and \textit{TrajFlow} are better calibrated when the task requires preserving sharper conditional dependencies between origins and downstream movement. 
On Chengdu, \textit{DiffTraj} has a slight overall advantage, especially in medium- and low-frequency origin regimes, whereas \textit{TrajFlow} performs best on the most frequent origins. Together with the supplementary results on Porto and Shanghai reported in Appendix~\ref{app:Additional Conditional Destination Results on Porto and Shanghai}, these findings suggest that \textit{DiffTraj} and \textit{TrajFlow} form the strongest pair for conditional mobility consistency, while their relative ranking remains dataset-dependent.
This suggests that continuous generative models without explicit road-network decoding constraints may currently offer greater flexibility for modeling sharp conditional destination distributions.

\subsection{Computational Cost and Efficiency}
Table~\ref{tab:porto_efficiency} compares parameter count, training cost, and generation efficiency on Porto, which serves as a representative large-scale benchmark setting in our study. These results show that model quality should be interpreted together with computational practicality rather than in isolation. 
All efficiency numbers are measured under the benchmark execution setting used by each reproduced method on a given dataset and should be interpreted as comparative benchmark-side end-to-end cost rather than as hardware-independent absolute deployment cost or a strictly hardware-homogeneous throughput comparison.

\begin{table*}[t]
\centering
\caption{Comparison of computational cost and efficiency for different trajectory generation models on Porto.}
\label{tab:porto_efficiency}
\resizebox{\textwidth}{!}{
\begin{tabular}{lcccc|cccc}
\toprule
\multirow{2}{*}{Model}
& \multirow{2}{*}{Trainable Params}
& \multicolumn{3}{c|}{Training Cost}
& \multicolumn{4}{c}{Generation Cost} \\
\cline{3-9}
&
& Total Time & Avg. Epoch Time & Peak GPU Mem.
& Total Time & Avg. Time / Traj. & Throughput & Peak GPU Mem. \\
&
& (h / min / s) & (s / min) & (MB / GB)
& (h / min / s) & (ms) & (traj/s) & (MB / GB) \\
\midrule
Markov & N/A & 2.23 min & N/A & N/A & 5.11 h & 227.58 & 4.41 & N/A \\
VAE & 445.14 K & 21.08 min & 2.53 s & 64.43 MB & 1.23 s & 0.02 & 66001.24 & 13.20 MB \\
TrajVAE & 414.40 K & 18.12 h & 2.17 min & 7.11 GB & 31.84 s & 0.39 & 2539.31 & 13.94 MB \\
TrajGAN & 416.75 M & 5.48 h & 1.03 min & 7.78 GB & 5.24 s & 0.07 & 15409.27 & 1.58 GB \\
DiffTraj & 16.86 M & 11.34 h & 3.40 min & 16.41 GB & 18.74 min & 13.92 & 71.85 & 6.75 GB \\
DiffRNTraj & 27.18 M & 3.63 h & 1.68 min & 4.40 GB & 40.51 min & 30.09 & 33.23 & 13.82 GB \\
TrajFlow & 15.33 M & 6.22 h & 1.87 min & 4.29 GB & 8.96 min & 6.66 & 150.24 & 282.90 MB \\
\bottomrule
\end{tabular}
}
\end{table*}

At one end of the spectrum, \textit{VAE} is extremely efficient. It has only 445.14K trainable parameters, very low memory usage, and by far the highest generation throughput. However, this efficiency comes with a clear loss in benchmark fidelity. Although \textit{Markov} incurs almost no trainable optimization cost, its measured generation throughput is low under the present benchmark execution setting, since it is implemented as a CPU-side sampling procedure rather than GPU-batched neural inference. Accordingly, this number should be interpreted as benchmark-side end-to-end execution cost under the model's native implementation regime, not as a strictly hardware-homogeneous throughput comparison. From a deployment perspective, it nevertheless indicates that classical baselines may be less favorable when large-scale synthetic trajectory generation requires high sampling throughput.

Among the stronger models, \textit{DiffTraj} and \textit{DiffRNTraj} illustrate the typical cost of iterative diffusion-style generation. Both achieve strong fidelity, but generation is substantially slower than for the VAE-based models and also slower than for \textit{TrajFlow}. \textit{DiffTraj} has the largest training memory footprint, while \textit{DiffRNTraj} has the slowest generation throughput among the deep generative baselines. This confirms that gains in fidelity and structural realism can come with nontrivial deployment cost.

\textit{TrajFlow} is particularly notable from a practical perspective. It remains highly competitive on both global and trajectory-level quality, yet its generation cost is substantially lower than that of the diffusion models and its memory footprint is also moderate. This suggests that flow-matching-based generation may provide a promising compromise on the fidelity--efficiency frontier.

Taken together, Table~\ref{tab:porto_efficiency} indicates that the benchmark does not reveal a single universally optimal model, but rather a Pareto-like trade-off among fidelity, structural realism, and computational cost. In practice, the preferred model depends on whether the application prioritizes route-level fidelity, road conformity, or high-throughput generation.

\subsection{Discussion}

The experiments lead to three broader implications for urban trajectory generation.

First, benchmark conclusions depend strongly on \emph{which notion of realism is prioritized}. If the target application emphasizes fine-grained individual-trip fidelity, such as behavior replay or high-quality synthetic route generation, \textit{DiffTraj} is the most reliable choice in the current benchmark. If the application emphasizes city-scale spatial organization, road conformity, and hotspot recovery, \textit{DiffRNTraj} and \textit{TrajFlow} are especially attractive. If the goal is lightweight modeling of coarse endpoint statistics, \textit{Markov} remains a meaningful low-complexity reference.

Second, the best-performing methods should be understood as negotiating different trade-offs rather than solving the task in the same way. \textit{DiffTraj} is strongest in sample-level geometry, \textit{DiffRNTraj} is strongest in structure-aware realism, and \textit{TrajFlow} offers a particularly attractive balance between fidelity and efficiency. Shanghai is especially informative in this regard, because under more challenging data conditions, the tension between these objectives becomes sharper and model weaknesses become easier to expose.

Third, the results validate the central motivation of CityTrajBench. Without unified preprocessing, representation conversion, and multi-level evaluation, it would be easy to draw misleading conclusions from a narrow metric set. For example, one might conclude that \textit{Markov} is highly competitive if only endpoint statistics are considered, or that road-network-aware generation is always superior if only density realism is measured. By jointly evaluating global realism, trip-level fidelity, trajectory-level similarity, conditional consistency, and efficiency, CityTrajBench makes these trade-offs explicit and therefore provides a more reliable basis for future model development.

\section{Conclusion}

In this paper, we presented \emph{CityTrajBench}, a unified benchmark for city-scale vehicle trajectory generation. The benchmark standardizes data preprocessing, trajectory normalization, model adaptation, map-aware post-processing, model selection, and multi-level evaluation under a common experimental protocol, enabling fairer comparison across heterogeneous trajectory generators. By bringing classical statistical methods and modern deep generative models into the same evaluation framework, CityTrajBench addresses a long-standing challenge in trajectory generation research: the lack of reproducible and directly comparable benchmarking settings.

Using three real-world urban trajectory datasets and seven representative generation methods, we conducted a systematic empirical study from multiple perspectives, including global spatial realism, trip-level statistical fidelity, trajectory-level geometric similarity, conditional mobility consistency, and computational efficiency. The results show that trajectory generation quality is inherently multi-objective. Continuous generative models define the strongest frontier in the current benchmark, with \emph{DiffTraj} consistently achieving the best trajectory-level fidelity, \emph{DiffRNTraj} performing particularly well on structure-sensitive global metrics, and \emph{TrajFlow} offering a strong balance between realism and efficiency. At the same time, the benchmark also shows that simple statistical baselines such as \emph{Markov} remain competitive on certain coarse-grained endpoint-related statistics. These findings highlight that no single model dominates all aspects of urban trajectory generation equally, and that meaningful evaluation must consider both macro-level realism and micro-level plausibility.

Despite these contributions, the current benchmark has several limitations. First, CityTrajBench focuses on city-scale vehicle trajectory generation and therefore does not yet cover other important mobility modalities, such as pedestrian, cycling, or multimodal movement data. Second, although the benchmark includes heterogeneous model families and multiple real-world datasets, its current scope is still limited relative to the diversity of urban environments, sensing conditions, and sampling regimes encountered in practice. Third, the present evaluation protocol emphasizes spatial, statistical, and structural fidelity, but does not fully capture all aspects of temporal realism, behavioral causality, or downstream task utility. In particular, the benchmark adopts a fixed-length trajectory representation for comparability across heterogeneous models, which may smooth short trajectories and truncate long ones relative to their raw forms. In addition, benchmark-side representation unification includes lightweight map-aware projection before evaluation, so the reported realism metrics reflect post-processed common-format outputs rather than raw decoder outputs alone. Moreover, some model-specific engineering choices, such as feature access, output space, and checkpoint-selection behavior, are necessarily documented rather than fully homogenized, because enforcing a strict one-size-fits-all protocol would distort the original assumptions of several baselines. The conditional evaluation further shows that the strongest continuous generative models better preserve origin-dependent destination structure under the current benchmark setting, although their relative ranking varies across datasets. Finally, although the current benchmark includes conditional evaluation for all seven compared baselines under a unified origin-conditioned protocol, the current conditional setting is still limited relative to the broader space of controllable trajectory generation, such as jointly constrained origin--destination, departure time, travel purpose, and network restriction settings.

These limitations suggest several directions for future work. An important next step is to extend the benchmark to broader mobility settings, including multimodal trajectories, richer temporal annotations, and larger cross-city or cross-region datasets. It is also valuable to incorporate additional evaluation dimensions, such as privacy risk, fairness, robustness under distribution shift, and utility for downstream tasks including simulation, forecasting, and policy analysis. Another promising direction is to support more controllable generation settings, where user-defined constraints such as origin--destination pairs, departure time, travel purpose, or road restrictions can be assessed in a unified manner. Future benchmark extensions should also investigate variable-length evaluation, raw-output versus post-processed evaluation, and more explicit sensitivity analysis for benchmark-side normalization choices. Finally, we hope CityTrajBench can serve not only as an evaluation platform, but also as a foundation for reproducible research on urban mobility generation, helping the community develop stronger, more transparent, and more practically useful trajectory generation methods.

\bibliographystyle{IEEEtran}
\bibliography{sample-base}

@String{Computer = "{IEEE} Computer" }

@String{Springer = "Springer-Verlag" }

@article{kong2023mobility,
  title={Mobility trajectory generation: a survey},
  author={Kong, Xiangjie and Chen, Qiao and Hou, Mingliang and Wang, Hui and Xia, Feng},
  journal={Artificial Intelligence Review},
  volume={56},
  number={Suppl 3},
  pages={3057--3098},
  year={2023},
  publisher={Springer}
}

@article{xiong2023trajsgan,
  title={TrajSGAN: A semantic-guiding adversarial network for urban trajectory generation},
  author={Xiong, Gang and Li, Zhishuai and Zhao, Meihua and Zhang, Yu and Miao, Qinghai and Lv, Yisheng and Wang, Fei-Yue},
  journal={IEEE Transactions on Computational Social Systems},
  volume={11},
  number={2},
  pages={1733--1743},
  year={2023},
  publisher={IEEE}
}

@article{li2024urban,
  title={An urban trajectory data-driven approach for COVID-19 simulation},
  author={Li, Zhishuai and Xiong, Gang and Lv, Yisheng and Ye, Peijun and Liu, Xiaoli and Tarkoma, Sasu and Wang, Fei-Yue},
  journal={IEEE Transactions on Computational Social Systems},
  volume={11},
  number={3},
  pages={4290--4299},
  year={2024},
  publisher={IEEE}
}

@article{zhu2023difftraj,
  title={Difftraj: Generating gps trajectory with diffusion probabilistic model},
  author={Zhu, Yuanshao and Ye, Yongchao and Zhang, Shiyao and Zhao, Xiangyu and Yu, James},
  journal={Advances in Neural Information Processing Systems},
  volume={36},
  pages={65168--65188},
  year={2023}
}

@article{wei2024diff,
  title={Diff-rntraj: A structure-aware diffusion model for road network-constrained trajectory generation},
  author={Wei, Tonglong and Lin, Youfang and Guo, Shengnan and Lin, Yan and Huang, Yiheng and Xiang, Chenyang and Bai, Yuqing and Wan, Huaiyu},
  journal={IEEE Transactions on Knowledge and Data Engineering},
  volume={36},
  number={12},
  pages={7940--7953},
  year={2024},
  publisher={IEEE}
}

@inproceedings{liu2018trajgans,
  title={trajGANs: Using generative adversarial networks for geo-privacy protection of trajectory data (Vision paper)},
  author={Liu, Xi and Chen, Hanzhou and Andris, Clio},
  booktitle={Location privacy and security workshop},
  pages={1--7},
  year={2018}
}

@inproceedings{xu2021simulating,
  title={Simulating continuous-time human mobility trajectories},
  author={Xu, Nan and Trinh, Loc and Rambhatla, Sirisha and Zeng, Zhen and Chen, Jiahao and Assefa, Samuel and Liu, Yan},
  booktitle={Proc. 9th Int. Conf. Learn. Represent},
  pages={1--9},
  year={2021}
}

@inproceedings{yuan2022activity,
  title={Activity trajectory generation via modeling spatiotemporal dynamics},
  author={Yuan, Yuan and Ding, Jingtao and Wang, Huandong and Jin, Depeng and Li, Yong},
  booktitle={Proceedings of the 28th ACM SIGKDD Conference on Knowledge Discovery and Data Mining},
  pages={4752--4762},
  year={2022}
}

@inproceedings{zhu2024controltraj,
  title={Controltraj: Controllable trajectory generation with topology-constrained diffusion model},
  author={Zhu, Yuanshao and Yu, James Jianqiao and Zhao, Xiangyu and Liu, Qidong and Ye, Yongchao and Chen, Wei and Zhang, Zijian and Wei, Xuetao and Liang, Yuxuan},
  booktitle={Proceedings of the 30th ACM SIGKDD Conference on Knowledge Discovery and Data Mining},
  pages={4676--4687},
  year={2024}
}

@article{chen2021trajvae,
  title={Trajvae: A variational autoencoder model for trajectory generation},
  author={Chen, Xinyu and Xu, Jiajie and Zhou, Rui and Chen, Wei and Fang, Junhua and Liu, Chengfei},
  journal={Neurocomputing},
  volume={428},
  pages={332--339},
  year={2021},
  publisher={Elsevier}
}

@article{rao2020lstm,
  title={LSTM-TrajGAN: A deep learning approach to trajectory privacy protection},
  author={Rao, Jinmeng and Gao, Song and Kang, Yuhao and Huang, Qunying},
  journal={arXiv preprint arXiv:2006.10521},
  year={2020}
}

@article{kothari2021human,
  title={Human trajectory forecasting in crowds: A deep learning perspective},
  author={Kothari, Parth and Kreiss, Sven and Alahi, Alexandre},
  journal={IEEE Transactions on Intelligent Transportation Systems},
  volume={23},
  number={7},
  pages={7386--7400},
  year={2021},
  publisher={IEEE}
}

@inproceedings{feng2024unitraj,
  title={Unitraj: A unified framework for scalable vehicle trajectory prediction},
  author={Feng, Lan and Bahari, Mohammadhossein and Amor, Kaouther Messaoud Ben and Zablocki, {\'E}loi and Cord, Matthieu and Alahi, Alexandre},
  booktitle={European Conference on Computer Vision},
  pages={106--123},
  year={2024},
  organization={Springer}
}

@article{gonzalez2008understanding,
  title={Understanding individual human mobility patterns},
  author={Gonzalez, Marta C and Hidalgo, Cesar A and Barabasi, Albert-Laszlo},
  journal={nature},
  volume={453},
  number={7196},
  pages={779--782},
  year={2008},
  publisher={Nature Publishing Group UK London}
}

@inproceedings{gambs2012next,
  title={Next place prediction using mobility markov chains},
  author={Gambs, S{\'e}bastien and Killijian, Marc-Olivier and del Prado Cortez, Miguel N{\'u}{\~n}ez},
  booktitle={Proceedings of the first workshop on measurement, privacy, and mobility},
  pages={1--6},
  year={2012}
}

@article{choi2021trajgail,
  title={TrajGAIL: Generating urban vehicle trajectories using generative adversarial imitation learning},
  author={Choi, Seongjin and Kim, Jiwon and Yeo, Hwasoo},
  journal={Transportation Research Part C: Emerging Technologies},
  volume={128},
  pages={103091},
  year={2021},
  publisher={Elsevier}
}

@inproceedings{ouyang2018non,
  title={A non-parametric generative model for human trajectories.},
  author={Ouyang, Kun and Shokri, Reza and Rosenblum, David S and Yang, Wenzhuo},
  booktitle={IJCAI},
  volume={18},
  pages={3812--3817},
  year={2018}
}

@inproceedings{cao2021generating,
  title={Generating mobility trajectories with retained data utility},
  author={Cao, Chu and Li, Mo},
  booktitle={Proceedings of the 27th ACM SIGKDD conference on knowledge discovery \& data mining},
  pages={2610--2620},
  year={2021}
}

@inproceedings{li2026trajflow,
  title={TrajFlow: Nation-wide Pseudo GPS Trajectory Generation with Flow Matching Models},
  author={Li, Peiran and Wang, Jiawei and Zhang, Haoran and Shi, Xiaodan and Koshizuka, Noboru and Shimizu, Chihiro and Jiang, Renhe},
  booktitle={The Fourteenth International Conference on Learning Representations},
  year={2026}
}

@article{kingma2013auto,
  title={Auto-encoding variational bayes},
  author={Kingma, Diederik P and Welling, Max},
  journal={arXiv preprint arXiv:1312.6114},
  year={2013}
}

\appendices
\section{Additional Conditional Destination Results on Porto and Shanghai}
\label{app:Additional Conditional Destination Results on Porto and Shanghai}

In addition to the main Chengdu conditional evaluation reported in Sec.~\ref{sec:Conditional_Mobility_Consistency}, we provide origin-conditioned destination divergence results for Porto and Shanghai in Tables~\ref{tab:cond_dest_porto_appendix} and~\ref{tab:cond_dest_shanghai_appendix}. These supplementary results are included to assess whether the conditional benchmark findings generalize beyond the main dataset used for head/torso/tail visualization and discussion in the main text.

The appendix results show that the relative strengths of \textit{DiffTraj} and \textit{TrajFlow} vary across datasets, but both remain among the strongest models in conditional destination consistency under the unified protocol.

\begin{table}[h]
    \caption{Origin-conditioned destination divergence on Porto.}
    % \label{tab:Origin-conditioned destination divergence on Porto}
    \label{tab:cond_dest_porto_appendix}
    \centering
    \resizebox{\linewidth}{!}{%
        \begin{tabular}{lrrrrr}
        \toprule
             Model &  AvgCondDestError &   Head &  Torso &   Tail &  NumOrigins \\
        \midrule
            Markov &            0.5883 & 0.5082 & 0.5695 & 0.6317 &         210 \\
               VAE &            0.4059 & 0.2297 & 0.3394 & 0.5162 &         210 \\
           TrajVAE &            0.4590 & 0.3415 & 0.4274 & 0.5250 &         210 \\
           TrajGAN &            0.4321 & 0.2885 & 0.3762 & 0.5231 &         210 \\
          DiffTraj &            \underline{0.1961} & \underline{0.0461} & \underline{0.1328} & \underline{0.2942} &         210 \\
        DiffRNTraj &            0.3917 & 0.1842 & 0.3264 & 0.5140 &         210 \\
          TrajFlow &           \textbf{ 0.1771} & \textbf{0.0353} & \textbf{0.1134} & \textbf{0.2721} &         210 \\
        \bottomrule
        \end{tabular}
        }
\end{table}

\begin{table}[h]
    \caption{Origin-conditioned destination divergence on Shanghai.}
    % \label{tab:Origin-conditioned destination divergence on Porto}
    \label{tab:cond_dest_shanghai_appendix}
    \centering
    \resizebox{\linewidth}{!}{%
        \begin{tabular}{lrrrrr}
        \toprule
             Model &  AvgCondDestError &   Head &  Torso &   Tail &  NumOrigins \\
        \midrule
            Markov &            0.5444 & 0.4332 & 0.5157 & 0.6063 &         144 \\
               VAE &            0.5410 & 0.4202 & 0.5445 & 0.5875 &         144 \\
           TrajVAE &            0.6007 & 0.5191 & 0.5913 & 0.6392 &         144 \\
           TrajGAN &            0.5345 & 0.4134 & 0.5359 & 0.5823 &         144 \\
          DiffTraj &            \underline{0.4712} & \underline{0.3256} & \underline{0.4441} & \underline{0.5460} &         144 \\
        DiffRNTraj &            0.5052 & 0.3335 & 0.5018 & 0.5764 &         144 \\
          TrajFlow &            \textbf{0.3203} & \textbf{0.1709} & \textbf{0.2874} & \textbf{0.4001} &         144 \\
        \bottomrule
        \end{tabular}
        }
\end{table}

\section{Configuration summary of CityTrajBench}
\label{app:Configuration_summary}

For clarity and reproducibility, Table~\ref{tab:benchmark_config_summary} summarizes the exact benchmark configuration used in all reported experiments.

\begin{table}[t]
\centering
\caption{Configuration summary of CityTrajBench.}
\label{tab:benchmark_config_summary}
\scriptsize
\setlength{\tabcolsep}{2.5pt}
\renewcommand{\arraystretch}{0.98}

\begin{tabularx}{\columnwidth}{
>{\raggedright\arraybackslash}p{1.45cm}
>{\raggedright\arraybackslash}p{1.85cm}
>{\raggedright\arraybackslash}X}
\toprule
\textbf{Category} & \textbf{Item} & \textbf{Setting} \\
\midrule

\multirow{8}{*}{\makecell[l]{Dataset}}
& Split unit & Trajectory-level split by \texttt{moveid}. \\
& Split ratio & 70\% / 15\% / 15\%. \\
& Split rule & Unique \texttt{moveid}s are shuffled, then split by ratio. \\
& Seed & 42. \\
& Order & Intra-trajectory point order is preserved. \\
& Bounds & Porto: lat [41.1399, 41.1867], lng [-8.6888, -8.5557]; Chengdu: lat [30.6531, 30.7297], lng [104.0354, 104.1299]; Shanghai: lat [31.2, 31.25], lng [121.405, 121.48]. \\
& Grid size & \(64\times64\) for all datasets. \\
& Coord.\ norm. & Min--max normalized to \([-1,1]\) using each dataset bounding box; fixed length \(L=200\). \\
\midrule

\multirow{4}{*}{\makecell[l]{Length\\normalization\\\&Features}}
& Short traj. & Linear interpolation to length 200. \\
& Long traj. & Prefix truncation to the first 200 points. \\
& Cond.\ feat.\ dim. & 8 (departure-time slot, trip distance, trip duration, trajectory length, average step distance, average speed, start ID, end ID). \\
& Norm.\ stats & Fitted on train split; applied to val / test. \\
\midrule

\multirow{4}{*}{\makecell[l]{Dataset-\\specific\\preprocessing}}
& Interval & Unified to 30 seconds across datasets. \\
& Departure time & 288 five-minute bins when timestamps are valid; otherwise 0. \\
& Start / end IDs & Row-major IDs on a \(16\times16\) grid over the dataset bounding box. \\
& Trip distance & Sum of geodesic distances along the fixed-length trajectory. \\
\midrule

\multirow{5}{*}{\makecell[l]{Spatial\\metrics}}
& Density Error & JSD over normalized grid-cell density. \\
& Pattern Score & Top-\(K\) hotspot threshold with \(K=409\) (\(\approx 10\%\) of cells). \\
& Trip Error & Mean JSD of start and end distributions on the shared grid. \\
& Length Error & 20 bins from pooled real and generated trajectory lengths. \\
& JSD-SD & 20 bins from pooled real and generated step distances. \\
\midrule

\multirow{8}{*}{\makecell[l]{Traj.\\match}}
& Match space & Local planar \(xy\) coordinates in meters converted from latitude / longitude. \\
& KD-tree summary & Start \(x,y\), end \(x,y\), and total trajectory length. \\
& Summary dim. & 5. \\
& Feature norm. & Standardized by real-set mean and std. \\
& Retrieval & KD-tree nearest-neighbor query. \\
& Pool size & Top-50. \\
& Embedding & 32 resampled points. \\
& Re-ranking & Euclidean distance on the 32-point polyline embedding. \\
\midrule

\multirow{7}{*}{\makecell[l]{Conditional\\destination\\evaluation}}
& Eval.\ grid & \(16\times16\). \\
& Origin unit & Start-point grid cell. \\
& Dest.\ unit & End-point grid cell on the same grid. \\
& Cond.\ dist. & Normalized destination-cell histogram conditioned on an origin cell. \\
& Div.\ metric & JSD between real and generated conditional destination distributions. \\
& AvgCondDestError & Mean JSD over all valid origin cells. \\
& Valid origin & At least 5 real trajectories per origin cell. \\
\midrule

\multirow{4}{*}{\makecell[l]{Origin\\groups}}
& Ranking & Descending real-data start-cell frequency. \\
& Head & Top 20\% of valid origin cells. \\
& Torso & Middle 30\% of valid origin cells. \\
& Tail & Bottom 50\% of valid origin cells. \\
\midrule

\multirow{3}{*}{\makecell[l]{Conditional\\visualization}}
& Grid & \(16\times16\). \\
& Origin selection & One head / torso / tail origin cell selected from valid real origins. \\
& Valid threshold & At least 30 real trajectories per origin cell. \\
\midrule

\multirow{3}{*}{\makecell[l]{Post-\\proc.}}
& Coord.\ recovery & Inverse mapping from normalized space to dataset GPS bounds. \\
& DiffRNTraj & Treated as already in GPS / road-network space; no inverse normalization. \\
& Map matching & Point-wise nearest-road projection using an STRtree spatial index. \\
\midrule

\multirow{3}{*}{\makecell[l]{Runs}}
& Seeds & 42, 52, 62. \\
& Statistic & Mean \(\pm\) std across seeds. \\
& Sample size & Matched to the corresponding test split size. \\
\midrule

\multirow{3}{*}{\makecell[l]{Hardware}}
& GPU & NVIDIA A800 80GB PCIe. \\
& CUDA / driver & CUDA 12.4; NVIDIA driver 550.127.05. \\
& Runtime note & Markov runs on CPU for fitting and generation; neural models use GPU inference. \\
\bottomrule
\end{tabularx}

\end{table}

\clearpage
\clearpage

{
\vspace{-45pt}
\begin{IEEEbiography}[{\includegraphics[width=1in,height=1.25in,clip,keepaspectratio]{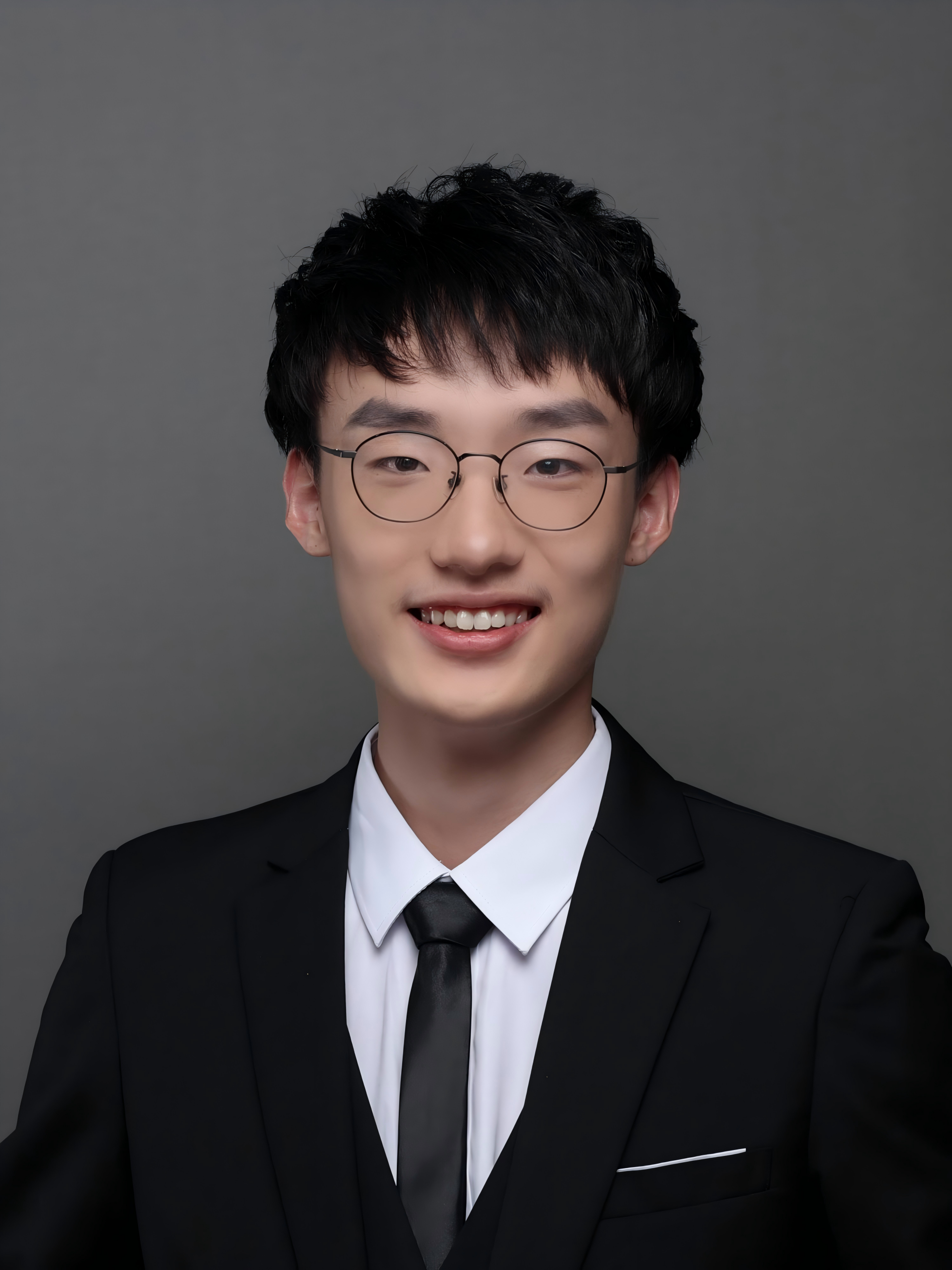}}]{Shibo Zhu} is currently pursuing his Ph.D. in Building Energy and Environment Engineering at The Hong Kong Polytechnic University. He holds a Bachelor's degree in Computer Science from the Southern University of Science and Technology. His research interests include time series prediction, federated learning, and reinforcement learning. Additionally, Zhu is exploring the application of computer science technologies in enhancing building energy efficiency.
\end{IEEEbiography}
\vspace{-45pt}
\begin{IEEEbiography}[{\includegraphics[width=1in,height=1.25in,clip,keepaspectratio]{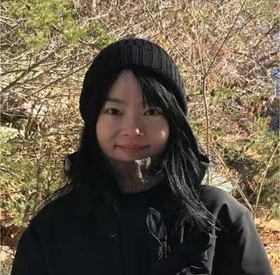}}]{Xiaodan Shi} received the B.E. and M.S. degrees in photogrammetry and remote sensing from Wuhan University, China. She received the Ph.D. degree with the Center for Spatial Information Science, The University of Tokyo, Kashiwa, Japan. She is working as an assistant professor with Department of Computer and Systems Sciences at Stockholm University. Her current research interests include people and vehicle trajectory prediction, trajectory generation and time series prediction for renewable energy.
\end{IEEEbiography}
\vspace{-45pt}
\begin{IEEEbiography}[{\includegraphics[width=1in,height=1.25in,clip,keepaspectratio]{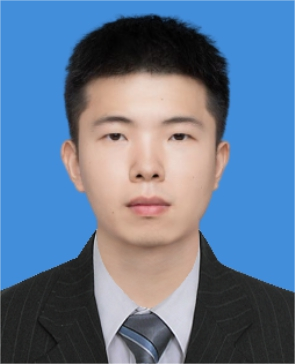}}]{Dayin Chen} is currently pursuing his Ph.D. in Building Energy and Environment Engineering at The Hong Kong Polytechnic University. He holds both a Bachelor's and a Master's degree in Computer Science from the Southern University of Science and Technology. His research interests include mobile computing, crowdsourcing, and neural architecture search. Additionally, Chen is exploring the application of computer science technologies in enhancing building energy efficiency.
\end{IEEEbiography}
\vspace{-45pt}
\begin{IEEEbiography}[{\includegraphics[width=1in,height=1.25in,clip,keepaspectratio]{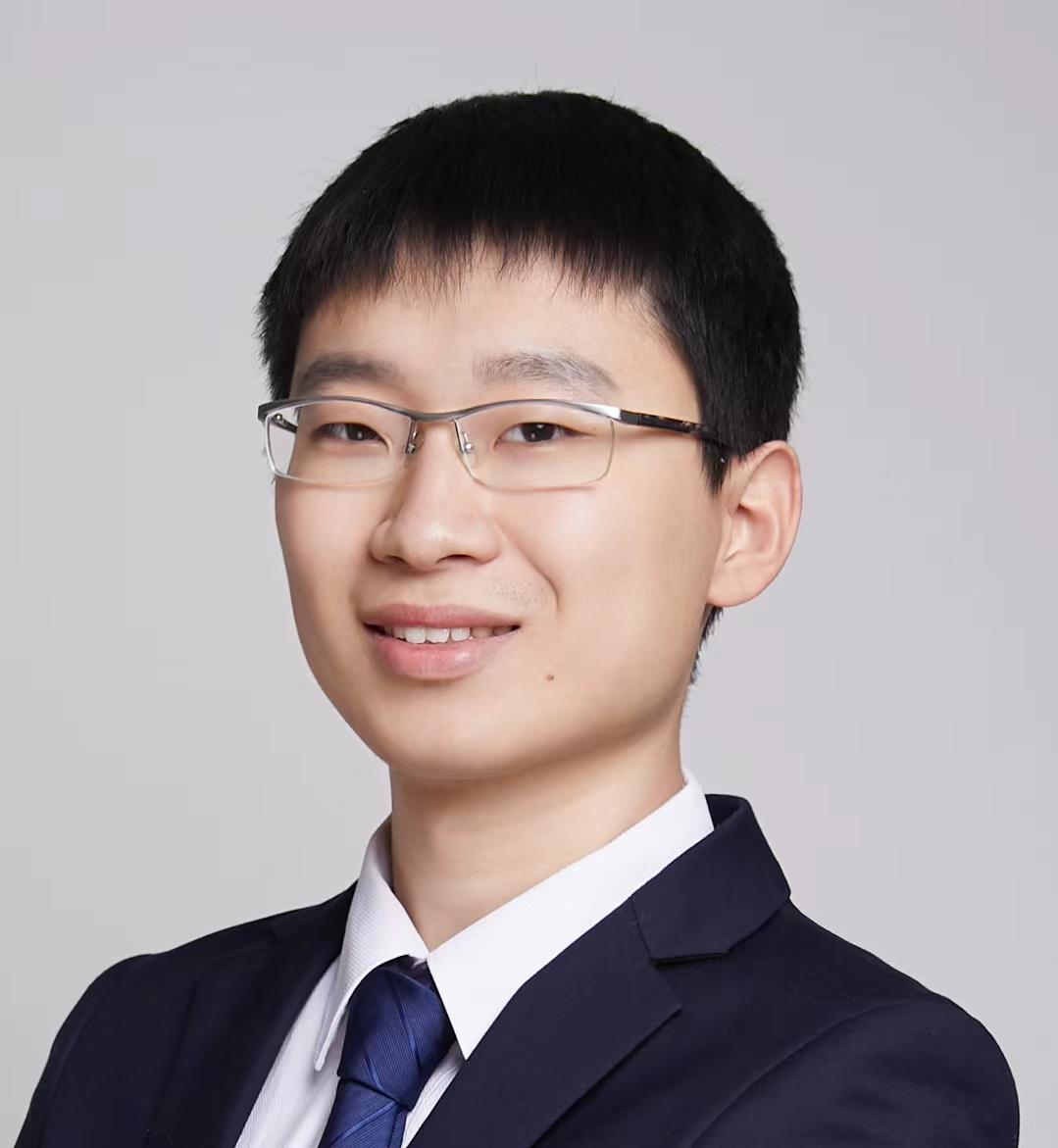}}]{Yuntian Chen} is an assistant professor at Eastern Institute of Technology, Ningbo. He received the B.S. degree from Tsinghua University, Beijing, China, in 2015, the dual B.S. degree from Peking University, Beijing, China, in 2015, and the Ph.D. degree with merit from Peking University, Beijing, China, in 2020. His research field includes scientific machine learning and intelligent energy systems. He is interested in the integration of domain knowledge and data-driven models.
\end{IEEEbiography}
\vspace{-45pt}
\begin{IEEEbiography}[{\includegraphics[width=1in,height=1.25in,clip,keepaspectratio]{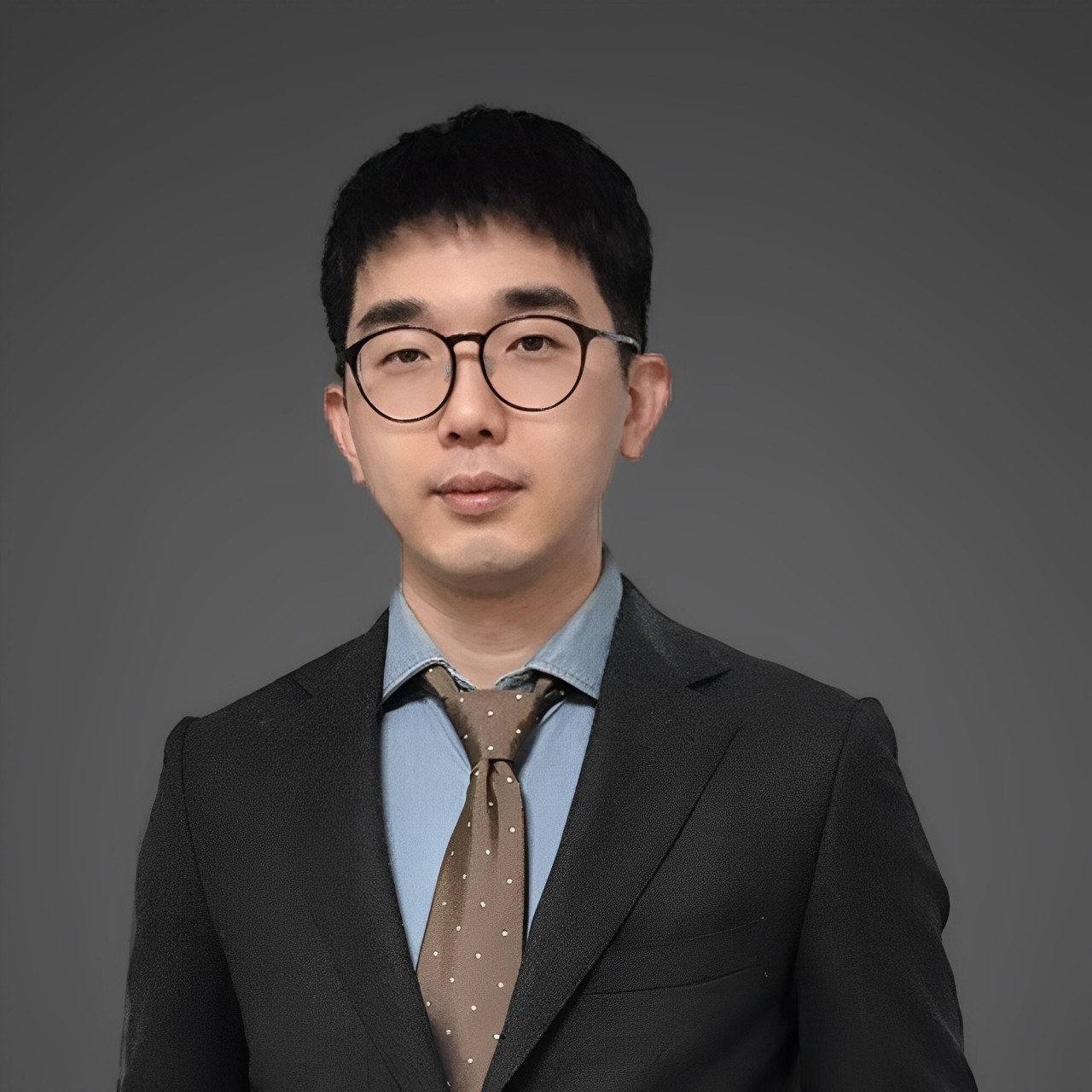}}]{Hao-Ran Zhang} is a Data Scientist at LocationMind Inc., Tokyo 101-0042, Japan. His current research and work focus on spatial data analytics, AI-driven human flow prediction, and the application of location-based big data in fields like urban traffic planning and disaster risk management. He has hands-on experience in developing machine learning models for processing and visualizing large-scale GPS-derived mobility data, and has contributed to optimizing LocationMind’s core services.
\end{IEEEbiography}
\vspace{-38pt}
\begin{IEEEbiography}[{\includegraphics[width=1in,height=1.25in,clip,keepaspectratio]{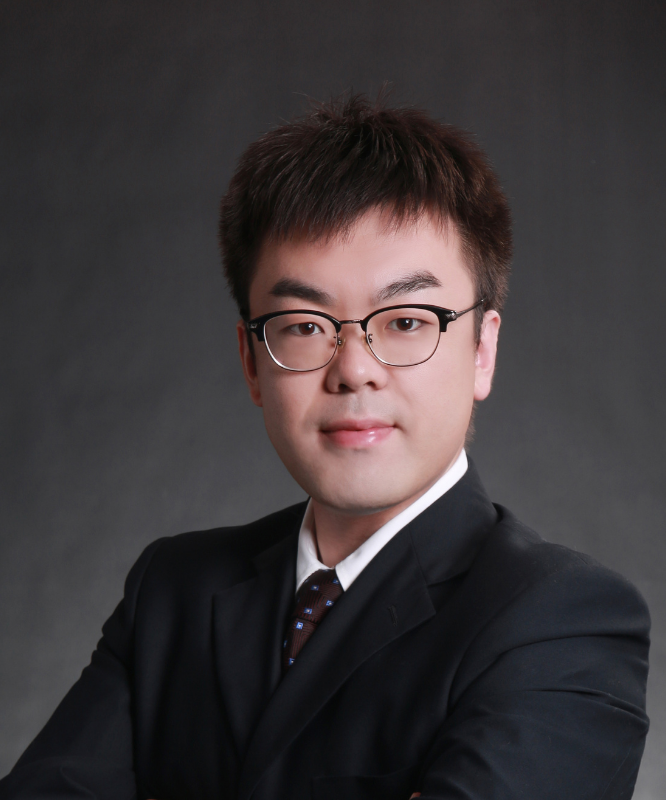}}]{Tianhao Wu} graduated from Peking University with dual Bachelor's degrees in Engineering and Economics, and a Ph.D. in Science. He has completed postdoctoral research at the U.S. Reservoir Engineering Institute and Rice University. As a key researcher, he has participated in major international projects funded by the National Natural Science Foundation of China and collaborated with companies like ExxonMobil and Saudi Aramco. His work has been featured as a cover article in Nano Energy and highlighted in Water Resources Research. He also reviews for several prominent journals, including SPE Journal, Fuel, Computational Geosciences, Energy \& Fuels, Journal of Petroleum Science and Engineering, and Journal of Natural Gas Science and Engineering.
\end{IEEEbiography}
\vspace{-25pt}
\begin{IEEEbiography}[{\includegraphics[width=1in,height=1.25in,clip,keepaspectratio]{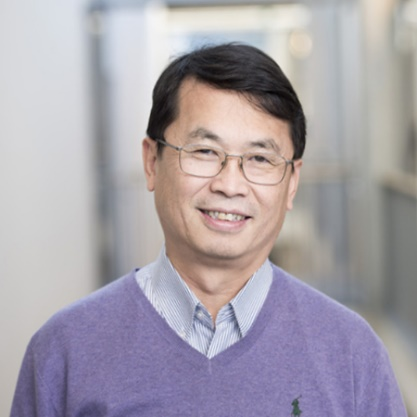}}]{Jerry Yan} is a member of the European Academy of Sciences and Arts and the Fellow of Hong Kong Academy of Engineering Sciences, currently serving as a Chair-Professor at the Hong Kong Polytechnic University. With a PhD from the Royal Institute of Technology (KTH), he has held chair professor positions at Lule{\aa} University of Technology, M\"alardalen University, and KTH, Sweden. Prof. Yan's research focuses on renewable energy, advanced energy systems, climate change mitigation, and environmental policies. He has a publication record of over 500 papers in renowned journals such as Science, Nature Energy, and Nature Climate Change, and holds more than 10 patents. Having supervised nearly 200 post-doctoral researchers and 50 doctoral candidates, Prof. Yan has secured substantial external grants exceeding 20 million Euros. Prof. Yan's contributions have been acknowledged through prestigious awards, including the Global Human Settlements Award of Green Technology, the EU Energy Islands' Award, Research2Business Top100, and the IAGE Lifetime Achievement Award.
\end{IEEEbiography}
}

\vfill

\end{document}